\PassOptionsToPackage{numbers}{natbib}
\documentclass[main]{article}

\usepackage[main, final]{neurips_2026}

\usepackage[utf8]{inputenc}
\usepackage[T1]{fontenc}
\usepackage{hyperref}
\usepackage{url}
\usepackage{booktabs}
\usepackage{amsfonts}
\usepackage{nicefrac}
\usepackage{microtype}
\usepackage{xcolor}

\usepackage[american]{babel}

\usepackage[dvipsnames, table]{xcolor}

\usepackage{hyperref}
\usepackage{url}
\usepackage{tikz,pgf}
\usepackage{stmaryrd}
\usepackage{graphicx}
\usepackage{wrapfig}
\usetikzlibrary{positioning,fit,shapes,patterns,arrows,arrows.meta,shapes,automata,calc,decorations.pathreplacing}
\usepackage{amsmath, amssymb, amsthm, dsfont, mathtools, multirow, mathrsfs, relsize,bm}

\usepackage[ruled,vlined]{algorithm2e}

\newtheorem{theorem}{Theorem}
\newtheorem{corollary}[theorem]{Corollary}
\newtheorem{lemma}[theorem]{Lemma}
\newtheorem{proposition}[theorem]{Proposition}

\newtheorem{definition}{Definition}

\newtheorem{remark}{Remark}
\theoremstyle{definition}
\newtheorem{example}{Example}

\usepackage{outlines}

\newcommand{\model}{\mathcal{M}}
\newcommand{\world}{\mathtt{w}}

\usepackage{tikz-cd}

\def\tikzcdequalsignoffset{0.1em}

\def\findedgesourcetarget#1#2{
    \let\sourcecoordinate\pgfutil@empty
    \ifx\tikzcd@startanchor\pgfutil@empty
        \def\tempa{\pgfpointanchor{#1}{center}}%
    \else
        \edef\tempa{\noexpand\pgfpointanchor{#1}{\expandafter\@gobble\tikzcd@startanchor}}
        \let\sourcecoordinate\tempa
    \fi
    \ifx\tikzcd@endanchor\pgfutil@empty
        \def\tempb{\pgfpointshapeborder{#2}{\tempa}}%
    \else
        \edef\tempb{\noexpand\pgfpointanchor{#2}{\expandafter\@gobble\tikzcd@endanchor}}%
    \fi
    \let\targetcoordinate\tempb
    \ifx\sourcecoordinate\pgfutil@empty%
        \def\sourcecoordinate{\pgfpointshapeborder{#1}{\tempb}}%
    \fi }
\tikzset{myequal/.style = {
    -, to path={\pgfextra{
        \findedgesourcetarget{\tikzcd@ar@start}{\tikzcd@ar@target}
        \ifx\tikzcd@startanchor\pgfutil@empty
            \def\tikzcd@startanchor{.center}
        \fi
        \ifx\tikzcd@endanchor\pgfutil@empty
            \def\tikzcd@endanchor{.center}
        \fi
        \pgfmathanglebetweenpoints{\pgfpointanchor{\tikzcd@ar@start}{\expandafter\@gobble\tikzcd@startanchor}}{\pgfpointanchor{\tikzcd@ar@target}{\expandafter\@gobble\tikzcd@endanchor}}
        \pgftransformrotate{\pgfmathresult}
        \pgfpathmoveto{\pgfpointadd{\sourcecoordinate}{\pgfpoint{0}{\tikzcdequalsignoffset}}}
        \pgfpathlineto{\pgfpointadd{\targetcoordinate}{\pgfpoint{0}{\tikzcdequalsignoffset}}}
        \pgfpathmoveto{\pgfpointadd{\sourcecoordinate}{\pgfpoint{0}{-\tikzcdequalsignoffset}}}
        \pgfpathlineto{\pgfpointadd{\targetcoordinate}{\pgfpoint{0}{-\tikzcdequalsignoffset}}}
        \pgfusepath{draw} }}}}

\usepackage{titletoc}
\newcommand\DoToC{%
  \startcontents
  \printcontents{}{1}{\textbf{Contents}\vskip3pt\hrule\vskip5pt}
  \vskip3pt\hrule\vskip5pt
}

\title{Tracing Actual Causes with\\Counterfactual Witness Maps}

\author{%
  Tim Woydt \\
  Department of Computer Science\\
  Technical University of Darmstadt\\
  \texttt{tim(dot)woydt(at)tu-darmstadt(dot)de} \\
  \And
  Matej Ze\v{c}evi\'{c} \\
  Department of Computer Science\\
  Technical University of Darmstadt\\
  \AND
  Kristian Kersting \\
  Department of Computer Science\\
  Technical University of Darmstadt\\
  Hessian Center for AI (hessian.AI)\\
  German Research Center for AI (DFKI)\\
  \And
  Moritz Willig \\
  Department of Computer Science\\
  Technical University of Darmstadt\\
}

\begin{document}
\maketitle

\begin{abstract}

Halpernian actual causation, the formal study of which events in a specific situation caused a specific outcome in acyclic structural causal models, underpins legal responsibility, moral blame, and the assessment of causal harm. Despite the rich logical framework, no prior sound and complete enumeration procedure is known for finding \emph{all} actual causes in a given setting, blocking downstream tasks such as responsibility attribution, blame assessment, and causal-harm grading that aggregate over the full cause set. We close this gap by introducing \emph{witness targets}: sets of joint configurations containing every counterfactual witness. In combination with an \emph{ancestor intervention grammar} we derive a finite directed acyclic \emph{witness map}. We prove that for every actual cause there exists a contingency set such that the joint intervention set appears as a node of this map, and present an algorithm for \emph{Tracing Actual Causes (TrAC)} that is sound and complete for enumerating all actual causes of any target event. We compare TrAC empirically with existing single-cause baselines on random Boolean structural causal models, demonstrating practical feasibility for graphs of up to 15 nodes.

\end{abstract}

\section{Introduction}\label{sec:intro}

What \emph{actually} caused a specific outcome, such as a patient's death after medication, an accident, or a software failure? Actual causation~\citep{Halpern2016} answers this question for singular, context-dependent cases, grounding legal responsibility~\citep{Moore2009}, moral blame~\citep{Cushman2008}, and algorithmic accountability~\citep{Chockler2004}. Downstream tasks that aggregate over the full cause set, notably responsibility and blame scores~\citep{Chockler2004} and the grading of causal harm~\citep{BeckersChocklerHalpern2022}, are blocked without an efficient enumeration procedure.
The dominant framework~\citep{HalpernPearl2005,Halpern2016} formalises actual causation in acyclic structural causal models (SCMs)~\citep{Pearl2009} via three conditions (AC1)--(AC3) on a candidate variable assignment, requiring its actual co-occurrence with the effect, a \emph{contingency set} of variables held at actual values exposing counterfactual dependence, and subset-minimality (formal definitions in Sec. \ref{sec:prelim}). We adopt the \emph{modified} HP definition~\citep{Halpern2015,Halpern2016}, one of several proposals in the literature, for its conceptual clarity and favourable complexity profile: verifying a single candidate is $\mathsf{D}_1^P$-complete~\citep{Aleksandrowicz2014,Halpern2015}, vs.\ $\Sigma_2^P$-complete in general under the original~\citep{Eiter2002}. The modified definition is also the natural fit for our construction: holding contingencies at their actual values is exactly the membership constraint of Corollary~\ref{cor:witness-existence}, so condition AC2 collapses to a single query against a precomputed target set, whereas the original definition's additional universal clause over contingency subsets admits no such collapse in general.

Existing exact algorithms suffer from two limitations: \citet{Ibrahim2020} translate HP conditions into ILP clauses for a \emph{single smallest} cause, restricted to Boolean SCMs and unable to enumerate the full cause set; \citet{Reyd2025} extend coverage to non-Boolean and black-box SCMs via forward simulation with sampled alternative values, but offer no soundness or completeness guarantee.
We present the first constructive, sound-and-complete algorithm for enumerating \emph{all} actual causes of an event in an acyclic SCM. Rather than iterating over alternative values forward, we propagate the \emph{complement of the target event} backward through the structural equations, yielding the \emph{witness targets}: for each candidate intervention set, the exact set of joint variable configurations compatible with that complement. This converts the existential search in AC2 into a single set-membership query, eliminating the dependence on alternative-value cardinality from the per-candidate verification cost. A finite \emph{witness map} over structurally admissible ancestor subsets then enumerates all candidate intervention sets; its worst-case cost remains exponential (Section~\ref{sec:algorithm}), which pruning substantially reduces in practice. Figure~\ref{fig:billy-suzy1} introduces the running rock-throwing preemption example (Suzy- and Billy throw rocks at a bottle with perfect accuracy (ST=1,BT=1): Suzy Hits (SH=1), but Billy does not hit (BH = 0) due to Suzy hitting first. The bottle shatters (BS=1) due to either of them hitting.) used throughout, with the formal machinery developed in Section~\ref{sec:finding-causes}.

\textbf{Contributions.}
We introduce three components that together yield a sound and complete enumeration procedure for actual causes. First, \emph{witness targets} (Section~\ref{sec:witness-targets}) provide an exact backward-propagation criterion for counterfactual satisfaction, converting AC2 into a set-membership query (Definitions~\ref{def:parent-targets}--\ref{def:global-targets}, Theorem~\ref{thm:witness-target}, Corollary~\ref{cor:witness-existence}). Second, the \emph{witness map} (Section~\ref{sec:witness-maps}) builds on the ancestor intervention grammar and ancestral map to enumerate all structurally admissible candidate intervention sets (Definitions~\ref{def:anc-grammar}--\ref{def:witness-map}, Propositions~\ref{prop:grammar-equiv}--\ref{prop:ancestor-reachability}). Third, the \emph{Tracing Actual Causes (TrAC)} algorithm (Section~\ref{sec:algorithm}, Algorithm~\ref{alg:find-causes}) combines both with soundness and completeness guarantees (Theorem~\ref{thm:sound-complete}) and an explicit complexity analysis (Appendix~\ref{app:complexity}).

\begin{figure*}[t]
\centering
\begin{minipage}[b]{0.59\textwidth}
\begin{tikzpicture}[
    dag node/.style={circle, draw, thick, minimum size=0.7cm, inner sep=1pt, font=\small},
    exo node/.style={dag node, dashed},
    lattice node/.style={rectangle, draw, thick, rounded corners=2pt, minimum width=0.9cm, minimum height=0.5cm, font=\scriptsize, inner sep=2pt},
    lbl/.style   = {draw = none, fill = none, font = \scriptsize,
                  inner sep = 1pt},
    arr/.style={-{Stealth[length=4pt]}, thick}, larr/.style={-{Stealth[length=3pt]}, thick, gray!70}, lab/.style={font=\tiny\itshape, text=black!90}, ]

\begin{scope}[xshift=0cm]
\node[font=\footnotesize\bfseries] at (0, 0) {(a)};
\node[dag node, fill=gray!15, double] (Y) at (0, 1) {$BS$};
\node[dag node, double] (S2) at (-1, 2.5) {$SH$};
\node[dag node, fill=red!10] (B2) at (1, 2.5) {$BH$};
\node[dag node, double] (S1) at (-1, 4) {$ST$};
\node[dag node, double] (B1) at (1, 4) {$BT$};
\node[exo node, double] (US) at (-1, 5.5) {\scriptsize$U_{ST}$};
\node[exo node, double] (UB) at (1, 5.5) {\scriptsize$U_{BT}$};
\draw[arr] (US)--(S1); \draw[arr] (UB)--(B1);
\draw[arr] (S1)--(S2); \draw[arr] (B1)--(B2); \draw[arr] (S2) to[bend left=45] (B2);
\draw[arr] (S2)--(Y); \draw[arr] (B2)--(Y);

\node[lbl, right=0.1cm of Y] {\scriptsize$=SH \!\vee\! BH$};
\node[lbl, right=0.1cm of S2] {\scriptsize$=ST$};
\node[lbl, right=0.1cm of B2] {\scriptsize$=\neg SH \!\wedge\! BT$};
\node[lbl, right=0.1cm of S1] {\scriptsize$=U_{ST}$};
\node[lbl, right=0.1cm of B1] {\scriptsize$=U_{BT}$};

\end{scope}
\begin{scope}[xshift=4.5cm]
\node[font=\footnotesize\bfseries] at (0, 0) {(b)};
\node[dag node, fill=gray!15] (L0) at (0, 1) {$BS$};
\node[lattice node] (L1S) at (-1.5, 3) {$SH$};
\node[lattice node, double, fill=red!10] (L1SB) at (0, 2.5) {$SH, BH$};
\node[lattice node] (L1B) at (1.5, 3) {$BH$};
\node[lattice node, double] (L2S1B2) at (.7, 4.15) {$SH, BT$};
\node[lattice node, double, fill=red!10] (L2S2B1) at (-0.7, 4.15) {$ST, BH$};
\node[lattice node] (L3S) at (-1.5, 5.5) {$ST$};
\node[lattice node, double] (L3SB) at (0, 5.5) {$ST, BT$};
\node[lattice node] (L3B) at (1.5, 5.5) {$BT$};
\draw[arr] (L0)--(L1S); \draw[arr] (L0)--(L1SB); \draw[arr] (L0)--(L1B);
\draw[arr] (L1S)--(L3S); \draw[arr] (L1B)--(L3B);
\draw[arr] (L1B)--(L1S); \draw[arr] (L1B)--(L2S1B2);
\draw[arr] (L1SB)--(L2S1B2); \draw[arr] (L1SB)--(L2S2B1);
\draw[arr] (L1SB)--(L1S); \draw[arr] (L1SB)--(L1B);
\draw[arr] (L2S1B2)--(L1S); \draw[arr] (L2S1B2)--(L3S); \draw[arr] (L2S1B2)--(L3SB); \draw[arr] (L2S1B2)--(L3B);
\draw[arr] (L2S2B1)--(L1B); \draw[arr] (L2S2B1)--(L3SB); \draw[arr] (L2S2B1)--(L3S); \draw[arr] (L2S2B1)--(L3B);
\draw[arr] (L3SB)--(L3S); \draw[arr] (L3SB)--(L3B);

\node[lbl, right=0.1cm of L0] {\scriptsize$\hat{\mathcal{O}}_{BS} = \{\neg BS\}$};
\end{scope}
\end{tikzpicture}
\end{minipage}%
\begin{minipage}[b]{0.1\textwidth}
    \smaller
    \begin{align*}
        &\hat{\mathcal{O}}_{do(ST,BT)} = \{(U_{ST}),(\neg U_{ST})\} \\
        &\hspace{25pt} \times \{(U_{BT}),(\neg U_{BT})\} \\
        &\hspace{25pt} \times \{(\neg ST,\neg BT, \neg SH,\neg BH,\neg BS)\} \\
        &\hat{\mathcal{O}}_{do(ST,\color{red}BH\color{black})} = \{(U_{ST}),(\neg U_{ST})\}  \\
        &\hspace{25pt} \times \{(U_{BT},BT),(\neg U_{BT},\neg BT)\} \\
        &\hspace{25pt} \times \{(\neg ST,\neg SH,\color{red}\neg BH\color{black},\neg BS)\} \\
        &\hat{\mathcal{O}}_{do(SH,BT)} = \{(U_{ST},ST),(\neg U_{ST},\neg ST)\}  \\
        &\hspace{25pt} \times \{(U_{BT}),(\neg U_{BT})\} \\
        &\hspace{25pt} \times \{(\neg BT,\neg SH,\neg BH,\neg BS)\} \\
        &\hat{\mathcal{O}}_{do(SH,\color{red}BH\color{black})} = \{(U_{ST},ST),(\neg U_{ST},\neg ST)\}  \\
        &\hspace{25pt} \times \{(U_{BT},BT),(\neg U_{BT},\neg BT)\} \\
        &\hspace{25pt} \times \{(\neg SH,\color{red}\neg BH\color{black},\neg BS)\}
    \end{align*} \vspace{1pt}
\end{minipage}
\caption{(a) The DAG of the Billy and Suzy example. The actual world is $(ST,BT,\neg BH,SH,BS)$ with environment $(U_{ST},U_{BT})$ (True variables are double-bordered). (b) The witness map (Definition~\ref{def:witness-map}) for the counterfactual target $\neg BS$ with double-bordered intervention sets for global targets (Definition~\ref{def:global-targets}) allowing the observed environment. \emph{As no singleton passes and none of these sets contains another, all four are but-for causes.} The nodes that allow for non-empty contingency ($\neg BH$) are highlighted in red. \emph{Thus we have two actual causes $ST(=\!1)$ and $SH(=\!1)$}. See Example~\ref{ex:billy-suzy} for details. (Best viewed in color.)}
\label{fig:billy-suzy1}
\end{figure*}

\section{Related Work}

\textbf{HP causation: extensions and algorithms.}
Foundational work formulates responsibility and blame~\citep{Chockler2004} (whose computation requires precisely the enumeration we provide), recasts HP via causal sufficiency~\citep{Beckers2021}, develops actual causation in CP-logic with graded judgements~\citep{BeckersVennekens2016}, and extends to probabilistic~\citep{FentonGlynn2017} and nondeterministic~\citep{Beckers2025} settings; downstream aggregations such as causal harm~\citep{BeckersChocklerHalpern2023quantifying} all presuppose an enumeration step. Context-specific independence (CSI)~\citep{Boutilier1996,Zhang1999} provides an algebraic lens on overdetermination and preemption that the outer approximation of Appendix~\ref{app:constraints} realises in marginal form~\citep{Pensar2015}; we discuss this connection further in Appendix~\ref{app:roadmap}.

\textbf{Explainability and attribution.}
Instance-level XAI sits close to but is distinct from actual causation. Shapley-value attributions, both associational~\citep{Lundberg2017,Janzing2020} and DAG-aware~\citep{Heskes2020}, quantify marginal contributions rather than identify singular causes. Counterfactual explanations~\citep{Wachter2018,Mothilal2020} and causally-augmented recourse~\citep{Karimi2022} implement but-for reasoning (AC2 without contingency), missing overdetermination and preemption. Population relative unit-level SCM explanations are discussed in~\citep{zevcevic2026causal}. \citet{Galhotra2021} apply probabilistic contrastive counterfactuals in databases; \citet{ChocklerHalpern2024image} bridge HP causation to image-level XAI under an assumed SCM, and causal root-cause analysis~\citep{Kim2013root,Budhathoki2022,Ikram2022root,Yang2022causal} addresses a related root-cause diagnosis problem under different assumptions and outside the HP framework; these methods neither target nor guarantee HP actual causes. We take the SCM as given and address the prerequisite enumeration step; SCM extraction is orthogonal.

\textbf{Adopted and new components.}
We adopt the modified HP definition (AC1)--(AC3)~\citep{Halpern2015,Halpern2016} with the responsibility measure used in the examples~\citep{Chockler2004,Halpern2016}, the complexity results for verifying a single candidate cause~\citep{Aleksandrowicz2014,Eiter2002}, which motivate our choice of definition, and the baselines with their solution strategies~\citep{Ibrahim2020,Reyd2025}, used unchanged for comparison; context-specific independence~\citep{Boutilier1996,Zhang1999,Pensar2015} serves only to interpret the pruning (Appendix~\ref{app:roadmap}). New are the witness targets (Definitions~\ref{def:parent-targets}--\ref{def:global-targets}) with their exactness (Theorem~\ref{thm:witness-target}) and the reduction of the AC2 test to a single set-membership query (Corollary~\ref{cor:witness-existence}); the ancestor intervention grammar, ancestral map and witness map (Definitions~\ref{def:anc-grammar}--\ref{def:witness-map}) with the closed-form child characterisation (Proposition~\ref{prop:grammar-equiv}) and the reachability criterion (Proposition~\ref{prop:ancestor-reachability}); TrAC with the first soundness-and-completeness guarantee for enumerating all actual causes (Theorem~\ref{thm:sound-complete}), where prior exact methods return a single smallest cause~\citep{Ibrahim2020} or lack guarantees~\citep{Reyd2025}; and the three-parameterisation complexity analysis (Appendix~\ref{app:complexity}), the constraint-based outer approximation (Appendix~\ref{app:constraints}) and the formally justified pruning strategies (Appendix~\ref{app:pruning}).

\section{Preliminaries}\label{sec:prelim}
\textbf{Structural causal models (SCM)} are structures $\model = \langle \mathbf{U}, \mathbf{V}, \mathcal{F}, \mathbb{P}_{\mathbf{U}} \rangle$ with a set of exogenous variables $\mathbf{U} = \{X_1,\dots,X_o\}$ and a set of endogenous variables $\mathbf{V} = \{X_{o+1},\dots,X_{o+p}\}$ with \emph{full configurations} $\mathbf{O} = \mathbf{U}\cup\mathbf{V}$ and values $\vec{o} \in \vec{\mathcal{O}} = \vec{\mathcal{U}} \times \vec{\mathcal{V}} = \bigtimes_{i=1}^{o+p} \mathcal{X}_i$, a set of structural equations $\mathcal{F} = \{X = F_{X} : X\in \mathbf{V} \}$ with functions $F_{X} :\vec{\mathcal{X}}_{pa(X)} \rightarrow \mathcal{X}_X$ and a distribution $\mathbb{P}_\mathbf{U}$ of the exogenous variables. Assignments\footnote{There is a canonical bijection between assignments $\vec{u}: \mathbf{U} \rightarrow \bigcup_{X_i \in \mathbf{U}} \mathcal{X}_i$ with $\vec{u}(X_i) \in \mathcal{X}_i$ and configurations $\vec{u} \in \vec{\mathcal{U}}$; we use both notations interchangeably and call either an \textit{environment}.} $\vec{u} : \mathbf{U} \rightarrow \bigcup_{X_i \in \mathbf{U}} \mathcal{X}_i$ of (all) exogenous variables are called \textbf{environments} of $\model$ and assignments $\world : \mathbf{V} \rightarrow \bigcup_{X_i \in \mathbf{V}} \mathcal{X}_i$ of (all) endogenous variables are called \textbf{worlds} of $\model$. A pair $(\model,\vec{u})$ is called a \textbf{setting} and the \textbf{actual world} $\world_{\vec{u}}$ of a specific setting $(\model, \vec{u})$ is induced by the evaluation of structural equations in $\mathcal{F}$. For an assignment $\vec{X} = \vec{x}$ with $\vec{X} \subseteq \mathbf{V}$, the \textbf{intervened SCM} $\model_{do(\vec{X}=\vec{x})}$ is obtained from $\model$ by replacing every structural equation $F_{X_i}$ with the constant $x_i$ (for $X_i \in \vec{X}$); its actual world under $\vec{u}$ is denoted $\world^*_{\vec{u}}$ and called \emph{counterfactual} when $\world^*_{\vec{u}} \neq \world_{\vec{u}}$. The corresponding \textbf{modal operator} satisfies $\model, \vec{u} \models [\vec{X}=\vec{x}]\varphi \;\Leftrightarrow\; \model_{do(\vec{X}=\vec{x})}, \vec{u} \models \varphi$.

    The actual causes easiest to define are so-called \textit{but-for causes}, which are \textit{necessary conditions} of an event. In a given setting $(\model, \vec{u})$ we say that the event $(\vec{X}=\vec{x})$ is a \textbf{but-for cause} of the event $\varphi$ if and only if $\model, \vec{u} \models (\vec{X}=\vec{x}) \wedge \varphi$ (AC1), there exists $\vec{x}' \in \vec{\mathcal{X}}$ with $\model, \vec{u} \models [\vec{X} = \vec{x}'] \neg\varphi$ (AC2$^*$), and $\vec{X}=\vec{x}$ is a smallest conjunction of atomic events satisfying the previous conditions (i.e., there is no subevent $(\vec{X}'=\vec{x}') \subset (\vec{X}=\vec{x})$ satisfying both AC1 and AC2$^*$) (AC3). Note here that a variable alone is never a cause; only the variable taking a specific value can be a cause. The atomic components $(X_i = x_i)$ for $X_i \in \vec{X}$ are referred to as \emph{parts} of the cause $(\vec{X} = \vec{x})$. To also capture causes whose but-for dependence is masked by a backup process, the modified HP definition~\citep{Halpern2015} relaxes the second condition as follows: if there exist $\vec{x}' \in \vec{\mathcal{X}}$ and $\vec{Z} \subseteq \mathbf{V} \setminus \vec{X}$ such that $\model, \vec{u} \models [\vec{Z}=\world_{\vec{u}}(\vec{Z}), \vec{X}=\vec{x}']\neg\varphi$ (AC2), and $\vec{X}=\vec{x}$ is a smallest conjunction of atomic events satisfying AC1 and AC2, then $\vec{X}=\vec{x}$ is called an \textbf{actual cause} of the event $\varphi$. Preemption is the canonical example (Figure~\ref{fig:billy-suzy1}, Appendix~\ref{app:example-billy-suzy}): Suzy's throw is not a but-for cause of the shattered bottle, but it is an actual cause, since holding Billy's miss at its actual value exposes the dependence.

    As the but-for causes and actual causes are defined via the existence of counterfactual worlds where the target event $\varphi$ does not hold, we call such a world $\world^*_{\vec{u}}$ a \textbf{counterfactual witness} for the actual cause or but-for cause $\vec{X}=\vec{x}$ of the event $\varphi$ in the setting $(\model, \vec{u})$. For a given setting $(\model, \vec{u})$, we can identify such a witness for an actual cause $\vec{X}=\world_{\vec{u}}(\vec{X})$ of an event $\varphi$ by a tuple $(\vec{Z},\vec{X},\vec{x}',\varphi)$ with $\vec{Z} \subseteq \mathbf{V} \setminus \vec{X}$. We call $\vec{Z}$ the \textbf{contingency set}: it consists of variables held fixed at their actual values $\world_{\vec{u}}(\vec{Z})$ in order to reveal the counterfactual dependence of $\varphi$ on $\vec{X}$. The causal claim $(\vec{X} = \world_{\vec{u}}(\vec{X}))$ thus holds as a cause \emph{contingent on} the context $\vec{Z} = \world_{\vec{u}}(\vec{Z})$. If $\vec{X}=\world_{\vec{u}}(\vec{X})$ is (also) a but-for cause of $\varphi$, then $\vec{Z} = \emptyset$, i.e., no contingency is required.

\section{Tracing Actual Causes}\label{sec:finding-causes}

We assume throughout that the SCM induces an acyclic causal graph.
W.l.o.g., the target $Y \in \mathbf{V}$ is a single variable with value
domain $\mathcal{Y}$; if not, introduce a fresh variable whose structural
equation copies the joint configuration of the target set. We write
$(Y \in \mathcal{Y}^*)$ with $\mathcal{Y}^* \subseteq \mathcal{Y}$ for the target event and $(Y \in \hat{\mathcal{Y}})$ for the induced counterfactual target with $\hat{\mathcal{Y}} = \mathcal{Y} \setminus \mathcal{Y}^*$.
We characterise the exact set of joint variable configurations compatible
with the counterfactual target via backward propagation through the structural equations. The construction proceeds through two building blocks: \emph{parent targets} and \emph{descendant targets}.

Our construction separates verification from enumeration. Section~\ref{sec:witness-targets} develops witness targets, computed by a single backward pass through the structural equations; they reduce the AC2 test for a fixed candidate pair $(\vec{X}, \vec{Z})$ to one membership query in a precomputed set (Corollary~\ref{cor:witness-existence}), with no search over alternative values. Section~\ref{sec:witness-maps} develops the ancestor intervention grammar and the witness map, a finite DAG whose nodes are the candidate intervention sets that have to be tested. Section~\ref{sec:algorithm} combines both into Algorithm~\ref{alg:find-causes} (TrAC) and proves it sound and complete (Theorem~\ref{thm:sound-complete}).

\subsection{Witness Targets}\label{sec:witness-targets}

\begin{definition}[parent targets]\label{def:parent-targets}
    For an endogenous variable $X \in \mathbf{V}$ and a set $\hat{\mathcal{O}} \subseteq \vec{\mathcal{O}}$, the \textbf{parent targets} are $\hat{\mathcal{O}}^{pa}_\model(X, \hat{\mathcal{O}}) := \bigl\{ \vec{o} \in \hat{\mathcal{O}} \bigm| \vec{o}_{pa(X)} \in F_X^{-1}(\vec{o}_X) \bigr\}$, where $\vec{o}_{pa(X)}$ is the projection of $\vec{o}$ onto the parent components of $X$ and $F_X^{-1}(\vec{o}_X) \subseteq \vec{\mathcal{X}}_{pa(X)}$ is the preimage of $\vec{o}_X$ under $F_X$.
\end{definition}

Note that the parent target operator is exact because a configuration $\vec{o}$ is retained if and only if its parent values are structurally consistent with producing $\vec{o}_X$, hence no feasible assignment is lost and no spurious one is introduced.

The parent target operator handles a single step of backward propagation. To accumulate constraints all the way from the target $Y$ up to any ancestor $X$, we apply it recursively across the graph.

\begin{definition}[descendant targets]\label{def:desc-targets}
    For a given SCM $\model$, target variable $Y \in \mathbf{V}$, target values $\hat{\mathcal{Y}} \subseteq \mathcal{Y}$, variable $X \in \mathbf{V}$, and intervention set $\vec{S} \subseteq \mathbf{V}$, the \textbf{descendant targets} $\hat{\mathcal{O}}^{\downarrow}_{\model,do(\vec{S})}(X, \hat{\mathcal{Y}})$ (written $\hat{\mathcal{O}}^{\downarrow}(X)$ when the context is fixed) are defined recursively by:
    \begin{align}\label{eq:desc-target}
        \hat{\mathcal{O}}^{\downarrow}_{\model,do(\vec{S})}(X, \hat{\mathcal{Y}}) :=
        \begin{cases}
            \bigl\{ \vec{o} \in \vec{\mathcal{O}} \;\big|\; \vec{o}_Y \in \hat{\mathcal{Y}} \bigr\} & \text{if } X = Y, \\
            \vec{\mathcal{O}} & \text{if } ch(X) = \emptyset, \\
            \displaystyle \bigcap\nolimits_{Z \in ch(X) \setminus \vec{S}} \!\!\!\!\!\quad   \hat{\mathcal{O}}^{pa}_\model\!\bigl(Z,\hat{\mathcal{O}}^{\downarrow}_{\model,do(\vec{S})}(Z,\hat{\mathcal{Y}})\bigr) & \text{otherwise.}
        \end{cases}
    \end{align}
    This is the exact set of joint configurations of $\model_{do(\vec{S})}$ compatible with $(Y \in \hat{\mathcal{Y}})$, as constrained by the causal mechanisms between $X$ and $Y$.
\end{definition}

The recursion traverses the causal graph from $Y$ upward. At the target $Y$, the set consists of all full configurations where $Y$ takes a value in $\hat{\mathcal{Y}}$. For leaves with no children, the set is unconstrained. Otherwise, the descendant targets of all non-intervened children are intersected via parent targets, propagating the joint feasibility constraint upward.

\begin{remark}[connection to causal sufficiency]\label{rem:beckers-sufficiency}
    The preimage $F_{X_i}^{-1}(\vec{o}_i)$ used by the parent target operator (Definition~\ref{def:parent-targets}) is exactly the set of joint parent configurations \emph{directly sufficient} for $\vec{o}_i$ in the sense of \citet{Beckers2021}; iterating across descendants (Definition~\ref{def:desc-targets}) realises a graph-aware version of \emph{strong sufficiency}, the transitive closure of direct sufficiency along the network.
\end{remark}

\begin{remark}[target reuse under intervention]\label{rem:reuse}
    For any variable $X \in \mathbf{V}$ that is either
    (i)~a strict descendant of some $S' \in \vec{S}$ (with $X \notin \vec{S}$),
    meaning $S'$ lies \emph{above} $X$ in the DAG and hence not on
    any directed path from $X$ toward $Y$, or
    (ii)~an element of $\vec{S}$ with $desc(X) \cap \vec{S} = \emptyset$,
    the intervention on $\vec{S}$ does not alter the structural equations
    on any directed path from $X$ to $Y$, so the descendant targets are invariant: $\hat{\mathcal{O}}^{\downarrow}_{\model,do(\vec{S})}(X,\hat{\mathcal{Y}}) = \hat{\mathcal{O}}^{\downarrow}_{\model,do(\emptyset)}(X,\hat{\mathcal{Y}})$.
    This enables efficient recomputation when moving along the witness map introduced in the next subsection; only the ancestors of and within the intervention set need to be recomputed; all descendant targets below the intervention are reused from the observational pass.
\end{remark}

The descendant targets $\hat{\mathcal{O}}^{\downarrow}_{\model,do(\vec{S})}(X, \hat{\mathcal{Y}})$ capture what is required at each individual variable $X$. To obtain the complete set of joint configurations compatible with the counterfactual target across the \emph{entire} model, we aggregate over all roots of the intervened model.

\begin{definition}[global targets]\label{def:global-targets}
    For a given SCM $\model$, target variable $Y \in \mathbf{V}$, target values $\hat{\mathcal{Y}} \subseteq \mathcal{Y}$, and intervention set $\vec{S} \subseteq \mathbf{V}$, the \textbf{global targets} $\hat{\mathcal{O}}_{\model,do(\vec{S})}(\hat{\mathcal{Y}}) := \bigcap_{R \in \mathbf{U} \cup \vec{S}} \hat{\mathcal{O}}^{\downarrow}_{\model,do(\vec{S})}(R,\hat{\mathcal{Y}})$ are the set of all variable configurations $\vec{o} \in \vec{\mathcal{O}}$ compatible with the target event $(Y \in \hat{\mathcal{Y}})$ when constraints are propagated backward from $Y$ through the structural equations of $\model_{do(\vec{S})}$\footnote{The global target is a superset of the actual solutions of $\model_{do(\vec{S})}$: it may contain configurations that disagree with $\model_{do(\vec{S})}$ on variables outside the ancestor cone of $Y$, since the backward propagation only constrains the path leading to $Y$. These spurious configurations are projected out by the environment-conditional membership query of Corollary~\ref{cor:witness-existence}.}, obtained by intersecting the descendant targets of all roots of $\model_{do(\vec{S})}$, that is, of all exogenous and all intervened variables. Concrete instances at the four pair-intervention sets of the Billy-and-Suzy example appear in the side panel of Figure~\ref{fig:billy-suzy1}.
\end{definition}

The global targets are constructed by a purely algebraic backward pass. The following theorem validates this construction for the variable configurations that are actually realised: for the unique joint configuration $\vec{o}_{\vec{u},\vec{s}}$ induced by $\model_{do(\vec{S}=\vec{s})}$ under environment $\vec{u}$, global-target membership is equivalent to counterfactual satisfaction of $(Y \in \hat{\mathcal{Y}})$.

\begin{theorem}[witness target membership]\label{thm:witness-target}
    Let $\model$ be an acyclic SCM, $Y \in \mathbf{V}$, $\hat{\mathcal{Y}} \subseteq \mathcal{Y}$, and $\vec{S} \subseteq \mathbf{V}$ with fixed values $\vec{s} \in \vec{\mathcal{S}}$. For the unique configuration
    $\vec{o}_{\vec{u},\vec{s}} := \bigl(\vec{u}(X_1),\dots,\vec{u}(X_o),$ $ \world^*_{\vec{u}}(X_{o+1}),\dots,\world^*_{\vec{u}}(X_{o+p})\bigr)$ of setting $(\model_{do(\vec{S}=\vec{s})},\vec{u})$ we have:
    \begin{align}\label{eq:witness-target-char}
        \vec{o}_{\vec{u},\vec{s}} \in \hat{\mathcal{O}}_{\model,do(\vec{S})}(\hat{\mathcal{Y}})
        \Leftrightarrow
        \model_{do(\vec{S} = \vec{s})}, \vec{u} \models (Y \in \hat{\mathcal{Y}}).
    \end{align}
\end{theorem}
    \textit{Proof sketch.}
    By induction on the reverse topological order of the DAG. The $(\Rightarrow)$ direction follows from the base case of Definition~\ref{def:desc-targets}, propagated along a directed path from a root of $\model_{do(\vec{S})}$ to $Y$. For $(\Leftarrow)$, acyclicity guarantees that $\vec{o}_{\vec{u},\vec{s}}$ is the unique configuration of $(\model_{do(\vec{S}=\vec{s})},\vec{u})$; membership in every descendant target set then follows by upward induction, using structural consistency of $F_Z$ at each step. The full inductive proof is given in Appendix~\ref{app:proofs}. \qed

Theorem~\ref{thm:witness-target} gives a membership criterion for \emph{fixed} values $\vec{s}$. For the AC2 condition, we need to check existence over \emph{all} alternative values $\vec{x}' \in \vec{\mathcal{X}}$. The following corollary shows that this existential quantifier can be collapsed into a single projection of the global targets, converting AC2 from a search into a membership test.

\begin{corollary}[witness existence]\label{cor:witness-existence}
    For an acyclic SCM $\model$, $Y \in \mathbf{V}$, $\hat{\mathcal{Y}} \subseteq \mathcal{Y}$, $\vec{X},\vec{Z} \subseteq \mathbf{V}$ with $\vec{X} \cap \vec{Z} = \emptyset$, there exists $\vec{o} \in \hat{\mathcal{O}}_{\model,do(\vec{X},\vec{Z})}(\hat{\mathcal{Y}})$ with $\vec{o}_\mathbf{U} = \vec{u}$ and $\vec{o}_{\vec{Z}} = \world_{\vec{u}}(\vec{Z})$, the actual values of $\vec{Z}$ (the restriction of the actual world $\world_{\vec{u}}$ to $\vec{Z}$), if and only if there exists some $\vec{x}' \in \vec{\mathcal{X}}$ such that $\model, \vec{u} \models [\vec{X}=\vec{x}',\vec{Z} = \world_{\vec{u}}(\vec{Z})](Y \in \hat{\mathcal{Y}})$. The membership condition thus collapses the existential quantifier over alternative values $\vec{x}' \in \vec{\mathcal{X}}$ in (AC2) into a set-membership query against the  global targets.
\end{corollary}
\textit{Proof.}
    $\vec{S} := \vec{Z} \cup \vec{X}$. $(\Leftarrow)$ Suppose there is $\vec{x}' \in \vec{\mathcal{X}}$ s.t. $\model, \vec{u} \models [\vec{X} = \vec{x}', \vec{Z} = \world_{\vec{u}}(\vec{Z})](Y \in \hat{\mathcal{Y}})$. By Theorem~\ref{thm:witness-target} (with intervention set $\vec{S}$ fixed at $(\vec{x}', \world_{\vec{u}}(\vec{Z}))$), the configuration $\vec{o}_{\vec{u},\vec{s}} \in \hat{\mathcal{O}}_{\model,do(\vec{S})}(\hat{\mathcal{Y}})$. Since $(\vec{o}_{\vec{u},\vec{s}})_\mathbf{U} = \vec{u}$ and $(\vec{o}_{\vec{u},\vec{s}})_{\vec{Z}} = \world_{\vec{u}}(\vec{Z})$, the configuration $\vec{o}_{\vec{u},\vec{s}}$ witnesses the existence statement. $(\Rightarrow)$ Suppose there exists $\vec{o}' \in \hat{\mathcal{O}}_{\model,do(\vec{S})}(\hat{\mathcal{Y}})$ with $\vec{o}'_\mathbf{U} = \vec{u}$ and $\vec{o}'_{\vec{Z}} = \world_{\vec{u}}(\vec{Z})$. Let $\vec{x}' := \vec{o}'_{\vec{X}}$. Since $\model$ is acyclic, the joint configuration of $\model_{do(\vec{X}=\vec{x}', \vec{Z}=\world_{\vec{u}}(\vec{Z}))}$ under environment $\vec{u}$ is \emph{unique} (each endogenous variable is determined by its parents and the structural equations); hence $\vec{o}' = \vec{o}_{\vec{u},\vec{x}',\world_{\vec{u}}(\vec{Z})}$. Theorem~\ref{thm:witness-target} then yields $\model, \vec{u} \models [\vec{X} = \vec{x}', \vec{Z} = \world_{\vec{u}}(\vec{Z})](Y \in \hat{\mathcal{Y}})$. \qed

Corollary~\ref{cor:witness-existence} solves the \emph{verification} half of the enumeration problem: given a candidate pair $(\vec{X}, \vec{Z})$, a single membership query suffices to decide AC2. What remains is the \emph{enumeration} half: which intervention sets $(\vec{X}, \vec{Z})$ need to be tested at all? Naively, one would search over all subsets of $\mathbf{V}$, a search space that is exponential in $|\mathbf{V}|$. The next subsection replaces this brute-force search with a grammar-based structure that generates only the \emph{structurally admissible} candidate sets via backward unfolding of the causal graph.

\subsection{Witness Maps}\label{sec:witness-maps}
\begin{wrapfigure}[19]{r}{0.42\textwidth}
\vspace{-3\baselineskip}
\centering
\begin{tikzpicture}[
    dag node/.style={circle, draw, thick, minimum size=0.65cm, inner sep=1pt, font=\small}, exo node/.style={dag node, dashed}, lattice node/.style={rectangle, draw, thick, rounded corners=2pt, minimum width=0.9cm, minimum height=0.5cm, font=\scriptsize, inner sep=2pt}, witness node/.style={rectangle, draw, thick, rounded corners=2pt, minimum width=1.6cm, minimum height=0.7cm, font=\scriptsize, inner sep=2pt, align=center}, arr/.style={-{Stealth[length=4pt]}, thick}, larr/.style={-{Stealth[length=3pt]}, thick, gray!70}, lab/.style={font=\tiny\itshape, text=black!90}, ]
\begin{scope}[xshift=0cm]
\node[font=\footnotesize\bfseries] at (0, 0) {(a)};
\node[dag node, fill=gray!15] (Y) at (0, .7) {$Y$};
\node[dag node] (C) at (.7, 1.8) {$C$};
\node[dag node] (B) at (.7, 3) {$B$};
\node[dag node] (A) at (-.7, 3) {$A$};
\node[exo node] (UA) at (0, 4) {\scriptsize$U_A$};
\draw[arr] (UA)--(A);
\draw[arr] (A)--(B); \draw[arr] (A)--(Y);
\draw[arr] (B)--(C); \draw[arr] (C)--(Y);
\end{scope}
\begin{scope}[xshift=2.5cm]
\node[font=\footnotesize\bfseries] at (0, 0) {(b)};
\node[dag node, fill=gray!15] (L0) at (0, .7) {$Y$};
\node[lattice node] (L1c) at (1.7, 2.5) {$C$};
\node[lattice node] (L1b) at (-1.1, 4) {$A$};
\node[lattice node] (L1d) at (.3, 1.9) {$A,C$};
\draw[arr] (L1d)--(L1b); \draw[arr] (L1d)--(L1c);
\draw[arr] (L0)--(L1b); \draw[arr] (L0)--(L1c); \draw[arr] (L0)--(L1d);
\node[lattice node] (L2c1) at (1.7, 4) {$B$};
\draw[arr] (L1c)--(L2c1); \draw[arr] (L2c1)--(L1b); \draw[arr] (L1d)--(L2c1);
\node[lattice node] (L2d4) at (.3, 3) {$A,B$};
\draw[arr] (L1d)--(L2d4); \draw[arr] (L2d4)--(L1b);
\draw[arr] (L2d4)--(L1b); \draw[arr] (L2d4)--(L2c1);
\end{scope}
\end{tikzpicture}
\caption{(a)~Example DAG with $pa(Y) = \{A,C\}$, $pa(C) = \{B\}$, $pa(B) = \{A\}$. (b)~Ancestral map $\mathcal{A}_\model(Y)$ (Definition~\ref{def:anc-map}): the root's children are the non-empty subsets of $pa^\mathbf{V}_\model(Y)$; the children of every other node follow from Proposition~\ref{prop:grammar-equiv}. Node $\{A,C\}$ has four children, $\{A\}$, $\{B\}$, $\{C\}$ and $\{A,B\}$ (Example~\ref{ex:derivation}).}
\label{fig:example-map}
\end{wrapfigure}
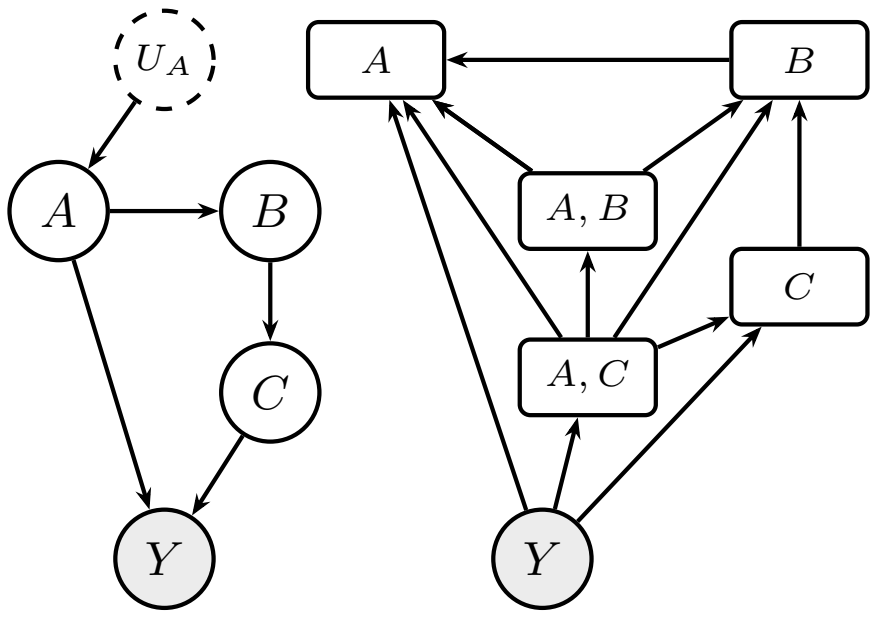
In the grammar, every variable of the current set either stays as it is (keep) or is replaced by its direct endogenous causes (unfold), moving the search one step further back through the graph. Starting from $\{Y\}$ and repeating these two local steps generates all candidate intervention sets; Figure~\ref{fig:example-map} shows a complete example.
\begin{definition}[ancestor intervention grammar]\label{def:anc-grammar}
    Let $\model$ have a DAG and $Y \in \mathbf{V}$. Write $\Sigma_\model(Y) := anc^\mathbf{V}_\model(Y) \cup \{Y\}$ for the endogenous ancestors of $Y$ including $Y$ itself. The \textbf{ancestor intervention grammar} $G_\model(Y)$ is the parallel set-rewriting system with \textbf{alphabet} $\Sigma_\model(Y)$, \textbf{start set} $\{Y\}$ and \textbf{productions} for every $X \in \Sigma_\model(Y)$:
    \begin{align}
        (P_{\mathrm{keep}})   &\quad X \longrightarrow \{X\},  \label{prod:keep}\\
        (P_{\mathrm{unfold}}) &\quad X \longrightarrow pa^\mathbf{V}_\model(X).
        \label{prod:unfold}
    \end{align}
    Both productions are applicable to every symbol (Remark~\ref{rem:productions} gives their causal reading); for $X$ with $pa^\mathbf{V}_\model(X) = \emptyset$, applying $(P_{\mathrm{unfold}})$ has the effect of \emph{removing} $X$ from the current set.

    A \textbf{derivation step} $\vec{S} \Rightarrow_{G} \vec{S}'$ between non-empty sets $\vec{S}, \vec{S}' \subseteq \Sigma_\model(Y)$ applies a production to each element independently. Choose $\vec{S} = \vec{S}_{\mathrm{unfold}} \;\dot{\cup}\; \vec{S}_{\mathrm{keep}}$ and set:
    \begin{align}\label{eq:deriv}
        \vec{S}' \subseteq\
        \bigcup_{X \in \vec{S}_{\mathrm{unfold}}} pa^\mathbf{V}_\model(X)
        \;\cup
        \vec{S}_{\mathrm{keep}},
        \qquad \vec{S}' \neq \emptyset,
    \end{align}
    where $\vec{S}_{\mathrm{unfold}} \neq \emptyset$ (at least one element is unfolded or dropped). Note that for $X \in \vec{S}_{\mathrm{unfold}}$ with $pa^\mathbf{V}_\model(X) = \emptyset$, applying $(P_{\mathrm{unfold}})$ contributes nothing to $\vec{S}'$, so unfolding a source node is equivalent to dropping it. We write $\vec{S} \Rightarrow^+_G \vec{S}'$ if $\vec{S}'$ is obtained from $\vec{S}$ by one or more derivation steps (the transitive closure of $\Rightarrow_G$). The \textbf{generated language} is $\mathcal{L}\bigl(G_\model(Y)\bigr) := \bigl\{\vec{S} \subseteq \Sigma_\model(Y) \bigm| \vec{S} \neq \emptyset,\, \{Y\} \Rightarrow^{+}_{G} \vec{S} \bigr\}$.
\end{definition}

\begin{remark}[productions and their roles]\label{rem:productions}
    The two productions have distinct causal meanings. \emph{Keep} ($X \in \vec{S}_{\mathrm{keep}}$): the variable is already at the right ancestral depth and will remain in the intervention set. \emph{Unfold} ($X \in \vec{S}_{\mathrm{unfold}}$): the variable's causal influence is traced one step further back through its endogenous parents $pa^\mathbf{V}_\model(X)$. This allows the grammar to generate combinations of ancestors at various depths, which is crucial for identifying actual causes that may lie at different levels of the causal graph. In particular, the grammar allows to move from an intervention set to any of its non-empty proper subsets in one step, since unfolded elements need not be selected for $\vec{S}'$.
\end{remark}

\begin{definition}[ancestral map]\label{def:anc-map}
    For an SCM $\model$ and target $Y \in \mathbf{V}$, the \textbf{ancestral map} $\mathcal{A}_\model(Y)$ is the directed graph whose nodes are the elements of $\mathcal{L}(G_\model(Y))$ together with the root $\{Y\}$, and whose edges are the derivation steps: $\vec{S} \to \vec{S}'$ in $\mathcal{A}_\model(Y)$ iff $\vec{S} \Rightarrow_G \vec{S}'$ in a single step.
\end{definition}

The ancestral map is defined implicitly via derivation steps of the grammar. For computational purposes, it is useful to have a closed-form expression for the children of any node $\vec{S}$ directly in terms of the graph structure of $\mathcal{G}_\model$, without simulating individual derivation steps.

\begin{proposition}[grammar–graph equivalence]\label{prop:grammar-equiv}
    For a set $\vec{S}'$, write $pa^\mathbf{V}_\model(\vec{S}') := \bigcup_{X \in \vec{S}'} pa^\mathbf{V}_\model(X)$, let $ch_{\mathcal{A}}(\vec{S})$ denote the set of children of a node $\vec{S}$ in $\mathcal{A}_\model(Y)$, and let $\mathcal{P}(\cdot)$ be the powerset operator. Then
    $ch_{\mathcal{A}}(\vec{S}) = \bigcup_{\emptyset \neq \vec{S}' \subseteq \vec{S}} \mathcal{P}\!\left( pa^\mathbf{V}_\model(\vec{S}') \cup (\vec{S} \setminus \vec{S}') \right) \setminus \{\emptyset\}$. The proof is given in Appendix~\ref{app:proofs}.
\end{proposition}

\begin{example}[derivation steps]\label{ex:derivation}
    In Figure~\ref{fig:example-map}, consider the node $\vec{S} = \{A, C\}$. Unfolding $C$ and keeping $A$ yields the candidate set $pa^\mathbf{V}_\model(C) \cup \{A\} = \{A, B\}$, whose non-empty subsets $\{A\}$, $\{B\}$ and $\{A,B\}$ are children. Unfolding $A$ drops it, since $pa^\mathbf{V}_\model(A) = \emptyset$, and keeping $C$ yields $\{C\}$; unfolding both yields $\{B\}$. Hence $ch_{\mathcal{A}}(\{A,C\}) = \{\{A\},\{B\},\{C\},\{A,B\}\}$, as stated by Proposition~\ref{prop:grammar-equiv}.
\end{example}

\begin{remark}[termination]\label{rem:termination}
    Since the DAG is acyclic and finite, every derivation from $\{Y\}$ terminates: each application of $(P_{\mathrm{unfold}})$ to $X$ replaces it by elements at strictly smaller topological depth, so $\mathcal{A}_\model(Y)$ is a finite directed acyclic graph. See Figure~\ref{fig:example-map} for an example.
\end{remark}

Finiteness bounds the search space; the next question is which ancestor sets actually appear as nodes. The following proposition gives a sufficient condition for reachability, a criterion we will invoke in the completeness proof of Theorem~\ref{thm:witness-suff} to show that every actual cause's intervention set is represented in the map.

\begin{proposition}[ancestor reachability]\label{prop:ancestor-reachability}
    Let $\vec{S} \subseteq anc^\mathbf{V}_\model(Y)$, $\vec{S} \neq \emptyset$. If every $X \in \vec{S}$ has a directed path to $Y$ in $\mathcal{G}_\model$ passing through no other element of $\vec{S}$, then $\vec{S} \in \mathcal{A}_\model(Y)$.
\end{proposition}

\textit{Proof.}
    Starting at the root node $\{Y\}$ in $\mathcal{A}_\model(Y)$, we iteratively unfold variables into their parents until some node contains an element $X \in \vec{S}$. We keep $X$ and continue unfolding the remaining variables until a node also contains the next element of $\vec{S}$, and so on. If some $X \in \vec{S}$ were not reachable by this procedure, every directed path from $X$ to $Y$ would pass through another element of $\vec{S}$ already included, contradicting the assumption.
\qed

\begin{remark}[sufficiency only]\label{rem:prop4-suffonly}
    The path-disjointness condition is sufficient but not necessary for $\vec{S} \in \mathcal{A}_\model(Y)$; Appendix~\ref{app:prop4-counterex} exhibits a reachable ancestor set on a chain-into-diamond DAG whose path-disjointness fails. The proposition is sufficient for the completeness argument of Theorem~\ref{thm:witness-suff}.
\end{remark}

We now have both ingredients needed to identify actual causes: the global targets from Section~\ref{sec:witness-targets}, which capture counterfactual satisfaction algebraically, and the ancestral map, which enumerates all structurally admissible candidate intervention sets. Combining them yields the \emph{witness map}: the ancestral map annotated at each node with the corresponding global targets.

\begin{definition}[witness map]\label{def:witness-map}
    For an ancestral map $\mathcal{A}_\model(Y)$ and target values $\hat{\mathcal{Y}} \subseteq \mathcal{Y}$, the \textbf{witness map} $\mathcal{W}_{\model}(Y \in \hat{\mathcal{Y}})$ is the directed graph obtained from $\mathcal{A}_\model(Y)$ by labelling each node $\vec{S}$ (a set of ancestors of $Y$) with $\hat{\mathcal{O}}_{\model,do(\vec{S})}(\hat{\mathcal{Y}})$, where the root node $\{Y\}$ is labelled $\hat{\mathcal{O}}_{\model,do(\emptyset)}(\hat{\mathcal{Y}})$. Each node thus encodes the exact set of joint variable configurations of $\model_{do(\vec{S})}$ compatible with the target event, and traversal of the map corresponds to systematically extending the intervention set along the causal ancestry of $Y$. Figure~\ref{fig:billy-suzy1}(b) displays the witness map for the Billy-and-Suzy preemption example.
\end{definition}

We now establish soundness (Lemma~\ref{lem:ac-witness}) and sufficiency of the witness map (Theorem~\ref{thm:witness-suff}), the latter directly inducing completeness of Algorithm~\ref{alg:find-causes}.

\begin{lemma}[AC-witness identification]\label{lem:ac-witness}
    Let $(\model, \vec{u})$ be a setting of an acyclic SCM with $\model, \vec{u} \models (Y \in \mathcal{Y}^*)$, and let $Y \in \mathbf{V}$, $\mathcal{Y}^* \subseteq \mathcal{Y}$, $\hat{\mathcal{Y}} := \mathcal{Y} \setminus \mathcal{Y}^*$. For disjoint $\vec{X}, \vec{Z} \subseteq \mathbf{V}$ the following two statements hold:
     \textbf{(But-for causes.)} If $\vec{X} \in \mathcal{L}(G_\model(Y))$ is a non-root node of $\mathcal{W}_\model(Y \in \hat{\mathcal{Y}})$ and there exists $\vec{o} \in \hat{\mathcal{O}}_{\model,do(\vec{X})}(\hat{\mathcal{Y}})$ with $\vec{o}_\mathbf{U} = \vec{u}$, then there exists $\vec{x}' \in \vec{\mathcal{X}}$ with $\vec{x}' \neq \world_{\vec{u}}(\vec{X})$ such that $(\emptyset, \vec{X}, \vec{x}', \mathcal{Y}^*)$ is a counterfactual witness for $(\vec{X} = \world_{\vec{u}}(\vec{X}))$ satisfying (AC1) and (AC2$^*$). In particular, $(\vec{X} = \world_{\vec{u}}(\vec{X}))$ is a but-for cause of $(Y \in \mathcal{Y}^*)$ if additionally $\vec{X}$ is minimal.
    \textbf{(Actual causes.)} If $\vec{X} \cup \vec{Z} \in \mathcal{L}(G_\model(Y))$ is a non-root node of $\mathcal{W}_\model(Y \in \hat{\mathcal{Y}})$ and there exists $\vec{o} \in \hat{\mathcal{O}}_{\model,do(\vec{X},\vec{Z})}(\hat{\mathcal{Y}})$ with $\vec{o}_\mathbf{U} = \vec{u}$ and $\vec{o}_{\vec{Z}} = \world_{\vec{u}}(\vec{Z})$, then there exists $\vec{x}' \in \vec{\mathcal{X}}$ with $\vec{x}' \neq \world_{\vec{u}}(\vec{X})$ such that $(\vec{Z}, \vec{X}, \vec{x}', \mathcal{Y}^*)$ is a counterfactual witness for $(\vec{X} = \world_{\vec{u}}(\vec{X}))$ satisfying (AC1) and (AC2) with contingency set $\vec{Z}$. In particular, $(\vec{X} = \world_{\vec{u}}(\vec{X}))$ is an actual cause of $(Y \in \mathcal{Y}^*)$ if additionally $\vec{X}$ is minimal.
\end{lemma}

\textit{Proof sketch.}
    (AC1) holds trivially since $\world_{\vec{u}}$ is the actual world. For (AC2), Corollary~\ref{cor:witness-existence} yields an $\vec{x}' \in \vec{\mathcal{X}}$ with $\model, \vec{u} \models [\vec{X}=\vec{x}', \vec{Z}=\world_{\vec{u}}(\vec{Z})](Y \in \hat{\mathcal{Y}})$; that $\vec{x}' \neq \world_{\vec{u}}(\vec{X})$ follows from the consistency of structural equations (setting $\vec{x}' = \world_{\vec{u}}(\vec{X})$ would reproduce the actual world, contradicting $Y \in \hat{\mathcal{Y}}$). See Appendix~\ref{app:proofs} for the full proof.\qed

Lemma~\ref{lem:ac-witness} handles the \emph{soundness} direction: any node in the witness map that passes the membership test certifies a genuine cause. The converse, which states that no actual cause can be missed, requires showing that every actual cause is \emph{represented} by some node in the map.

\begin{theorem}[sufficiency of witness map]\label{thm:witness-suff}
    Let $(\model, \vec{u})$ be a setting of an acyclic SCM with $\model, \vec{u} \models (Y \in \mathcal{Y}^*)$, and let $\hat{\mathcal{Y}} := \mathcal{Y} \setminus \mathcal{Y}^*$. For every actual cause $(\vec{X} = \world_{\vec{u}}(\vec{X}))$ of $(Y \in \mathcal{Y}^*)$, there exists a contingency set $\vec{Z} \subseteq \mathbf{V} \setminus \vec{X}$ such that $\vec{X} \cup \vec{Z} \in \mathcal{L}(G_\model(Y))$ and
    \begin{align}
        \exists\,\vec{o} \in \hat{\mathcal{O}}_{\model,do(\vec{X},\vec{Z})}(\hat{\mathcal{Y}}) \quad \text{with} \quad \vec{o}_\mathbf{U} = \vec{u} \;\;\text{and}\;\; \vec{o}_{\vec{Z}} = \world_{\vec{u}}(\vec{Z}).
    \end{align}
    In particular, but-for causes are the special case $\vec{Z} = \emptyset$, for which the membership condition reduces to the existence of $\vec{o} \in \hat{\mathcal{O}}_{\model,do(\vec{X})}(\hat{\mathcal{Y}})$ with $\vec{o}_\mathbf{U} = \vec{u}$.
\end{theorem}

\textit{Proof.}
    By (AC2), there exist $\vec{Z}_0 \subseteq \mathbf{V} \setminus \vec{X}$ and $\vec{x}' \neq \world_{\vec{u}}(\vec{X})$ such that $\model, \vec{u} \models [\vec{X} = \vec{x}',\, \vec{Z}_0 = \world_{\vec{u}}(\vec{Z}_0)](Y \in \hat{\mathcal{Y}})$. (For but-for causes, (AC2$^*$) gives this with $\vec{Z}_0 = \emptyset$ directly.) Let $\vec{Z} \subseteq \vec{Z}_0$ be inclusion-minimal such that $\model, \vec{u} \models [\vec{X} = \vec{x}',\, \vec{Z} = \world_{\vec{u}}(\vec{Z})](Y \in \hat{\mathcal{Y}})$ still holds; such a minimum exists since the collection of valid $\vec{Z}' \subseteq \vec{Z}_0$ is non-empty and finite. For \textbf{node existence: $\vec{X} \cup \vec{Z} \in \mathcal{L}(G_\model(Y))$.} By Proposition~\ref{prop:ancestor-reachability}, it suffices to show that every element of $\vec{X} \cup \vec{Z}$ has a directed path to $Y$ in $\mathcal{G}_\model$ passing through no other element of $\vec{X} \cup \vec{Z}$. Suppose for contradiction that some $V' \in \vec{X} \cup \vec{Z}$ has all directed paths to $Y$ blocked at some $V'' \in (\vec{X} \cup \vec{Z}) \setminus \{V'\}$. In $\model_{do(\vec{X}=\vec{x}',\, \vec{Z}=\world_{\vec{u}}(\vec{Z}))}$, the structural equation for $V''$ is overridden, so $V'$ cannot influence $\world^*_{\vec{u}}(Y)$ through $V''$. Since every causal path from $V'$ to $Y$ passes through $V''$, the value of $V'$ is causally inert for $Y$ in this model; hence $\model, \vec{u} \models [\vec{X} = \vec{x}',\, (\vec{Z} \cup \vec{X} \setminus \{V'\}) = \world_{\vec{u}}(\vec{Z} \cup \vec{X} \setminus \{V'\})](Y \in \hat{\mathcal{Y}})$. If $V' \in \vec{X}$, this contradicts the minimality (AC3) of $\vec{X}$ (we could drop $V'$ from the cause). If $V' \in \vec{Z}$, this contradicts the minimality of $\vec{Z}$. Hence every $V' \in \vec{X} \cup \vec{Z}$ has a path to $Y$ through no other element of $\vec{X} \cup \vec{Z}$, and $\vec{X} \cup \vec{Z} \in \mathcal{L}(G_\model(Y))$ follows. The \textbf{membership condition} now directly follows from Corollary~\ref{cor:witness-existence}.
\qed

\subsection{Algorithm}\label{sec:algorithm}

We summarise the identification procedure in Algorithm~\ref{alg:find-causes}. By Theorem~\ref{thm:witness-suff}, every actual cause admits a witness $(\vec{Z}, \vec{X}, \vec{x}', \mathcal{Y}^*)$ with $\vec{X} \cup \vec{Z} \in \mathcal{L}(G_\model(Y))$, so every relevant intervention set already appears as a node of the witness map.
By Corollary~\ref{cor:witness-existence}, the AC2 condition for a pair $(\vec{X}, \vec{Z})$ then reduces to a single membership query against the precomputed global targets $\hat{\mathcal{O}}_{\model,do(\vec{X} \cup \vec{Z})}(\hat{\mathcal{Y}})$.
The algorithm therefore treats each node $\vec{S} \in \mathcal{A}_\model(Y)$ directly as the total intervention set and enumerates all partitions $\vec{S} = \vec{X} \cup \vec{Z}$, where $\vec{X}$ is the candidate cause and $\vec{Z} = \vec{S} \setminus \vec{X}$ is the contingency set, testing each against these targets (Phase~2).

\begin{theorem}[soundness and completeness]\label{thm:sound-complete}
    Let $(\model, \vec{u})$ be a setting of an acyclic SCM with $\model, \vec{u} \models (Y \in \mathcal{Y}^*)$, and let $\hat{\mathcal{Y}} := \mathcal{Y} \setminus \mathcal{Y}^*$. Algorithm~\ref{alg:find-causes} returns exactly the collection $\mathcal{C}$ of variable sets $\vec{X} \subseteq \mathbf{V}$ such that $(\vec{X} = \world_{\vec{u}}(\vec{X}))$ is an actual cause of $(Y \in \mathcal{Y}^*)$: $\vec{X} \in \mathcal{C}$ iff $(\vec{X} = \world_{\vec{u}}(\vec{X}))$ is an actual cause of $(Y \in \mathcal{Y}^*)$.
\end{theorem}

\textit{Proof.}
    \textbf{Soundness ($\Rightarrow$).} Suppose $\vec{X} \in \mathcal{C}$ upon termination of Algorithm~\ref{alg:find-causes}. By Phase~2, there exists a node $\vec{S} \in \mathcal{A}_\model(Y)$ such that $\vec{X} \subseteq \vec{S}$ and, with $\vec{Z} := \vec{S} \setminus \vec{X}$, the membership test $\exists\,\vec{o} \in \hat{\mathcal{O}}_{\model,do(\vec{S})}(\hat{\mathcal{Y}})$ with $\vec{o}_\mathbf{U} = \vec{u}, \vec{o}_{\vec{Z}} = \world_{\vec{u}}(\vec{Z})$ passed. Since $\vec{S} \in \mathcal{L}(G_\model(Y))$, $\vec{S}$ is a non-root node of $\mathcal{W}_\model(Y \in \hat{\mathcal{Y}})$. Lemma~\ref{lem:ac-witness} then gives the existence of $\vec{x}' \neq \world_{\vec{u}}(\vec{X})$ such that $(\vec{Z}, \vec{X}, \vec{x}', \mathcal{Y}^*)$ is a counterfactual witness satisfying (AC1) and (AC2). Phase~3 retains only minimal elements, so $\vec{X}$ satisfies (AC3). Hence $(\vec{X} = \world_{\vec{u}}(\vec{X}))$ is an actual cause.

    \textbf{Completeness ($\Leftarrow$).} Suppose $(\vec{X} = \world_{\vec{u}}(\vec{X}))$ is an actual cause of $(Y \in \mathcal{Y}^*)$. By Theorem~\ref{thm:witness-suff}, there exists $\vec{Z} \subseteq \Sigma_\model(Y) \setminus \vec{X}$ such that $\vec{S} := \vec{X} \cup \vec{Z} \in \mathcal{L}(G_\model(Y))$ and some $\vec{o} \in \hat{\mathcal{O}}_{\model,do(\vec{X},\vec{Z})}(\hat{\mathcal{Y}})$ satisfies $\vec{o}_\mathbf{U} = \vec{u}, \vec{o}_{\vec{Z}} = \world_{\vec{u}}(\vec{Z})$. Hence the membership test for the partition $(\vec{X},\,\vec{Z})$ of $\vec{S}$ succeeds in Phase~2, so $\vec{X}$ is added to $\mathcal{C}$. Since $(\vec{X} = \world_{\vec{u}}(\vec{X}))$ satisfies (AC3) by assumption, $\vec{X}$ is not removed in Phase~3. Therefore $\vec{X} \in \mathcal{C}$ upon termination.
\qed

\begin{wrapfigure}[40]{r}{0.5\textwidth}\vspace{-1.5\baselineskip}
\begin{algorithm}[H]
\caption{Tracing Actual Causes (TrAC)}\label{alg:find-causes} \KwIn{SCM $\model$, environment $\vec{u}$, target variable $Y$, target event $\mathcal{Y}^*$} \KwOut{Set of (nodes of) actual causes $\mathcal{C} \subseteq \mathcal{P}(\mathbf{V})$} $\hat{\mathcal{Y}} \leftarrow \mathcal{Y} \setminus \mathcal{Y}^*$\; \tcc{Phase 1: Precompute observational global targets (single bottom-up pass)} \ForEach{$X \in \mathbf{V} \cup \mathbf{U}$ in reverse topological order}{ Compute $\hat{\mathcal{O}}^{\downarrow}_{\model,do(\emptyset)}(X, \hat{\mathcal{Y}})$\; } $\hat{\mathcal{O}}_{\model,do(\emptyset)}(\hat{\mathcal{Y}}) \leftarrow \bigcap_{U \in \mathbf{U}} \hat{\mathcal{O}}^{\downarrow}_{\model,do(\emptyset)}(U, \hat{\mathcal{Y}})$\; $\mathcal{C} \leftarrow \emptyset$\; \tcc{Check that the actual world is incompatible with the counterfactual target} \If{$\exists\,\vec{o} \in \hat{\mathcal{O}}_{\model,do(\emptyset)}(\hat{\mathcal{Y}})$ with $\vec{o}_\mathbf{U} = \vec{u}$}{
    \Return{$\mathcal{C}$}
} \tcc{Phase 2: Traverse witness map} \ForEach{node $\vec{S}$ in $\mathcal{A}_\model(Y)$}{
    \tcc{Recompute targets above $\vec{S}$ only; reuse below (Remark~\ref{rem:reuse})}
    Compute $\hat{\mathcal{O}}_{\model,do(\vec{S})}(\hat{\mathcal{Y}})$\;
    \tcc{Enumerate all cause/contingency partitions of $\vec{S}$}
    \ForEach{non-empty $\vec{X} \subseteq \vec{S}$}{
        $\vec{Z} \leftarrow \vec{S} \setminus \vec{X}$\;
        \If{$\exists\,\vec{o} \in \hat{\mathcal{O}}_{\model,do(\vec{S})}(\hat{\mathcal{Y}})$ with $\vec{o}_\mathbf{U} = \vec{u}$ and $\vec{o}_{\vec{Z}} = \world_{\vec{u}}(\vec{Z})$}{
            $\mathcal{C} \leftarrow \mathcal{C} \cup \{\vec{X}\}$\; } } } \tcc{Phase 3: Apply minimality (AC3)} Remove each $\vec{X} \in \mathcal{C}$ with a strict subset in $\mathcal{C}$\; \Return{$\mathcal{C}$}
\end{algorithm}
\end{wrapfigure}
Relative to the theoretical framework of Section~\ref{sec:witness-maps}, correctness is established by Theorem~\ref{thm:sound-complete} below: soundness rests on Lemma~\ref{lem:ac-witness}, which shows that every positive membership test corresponds to a genuine counterfactual witness; completeness rests on Theorem~\ref{thm:witness-suff}, which guarantees that every actual cause is represented by some node in the witness map. Phase~3 compares the collected candidates pairwise under set inclusion and removes every $\vec{X}$ for which some strict subset is also in $\mathcal{C}$. This filter is exact: by Lemma~\ref{lem:ac-witness}, every removed $\vec{X}$ has a strict subset satisfying AC1 and AC2 and hence violates AC3, and by Theorem~\ref{thm:witness-suff}, every actual cause is in $\mathcal{C}$, so every non-minimal $\vec{X}$ contains one as a strict subset and is removed.

Having established that Algorithm~\ref{alg:find-causes} is both sound and complete, we briefly comment on its computational cost. Crucially, Corollary~\ref{cor:witness-existence} eliminates any enumeration over alternative values $\vec{x}'$, so the cost depends only on the causal graph and intervention-set sizes, not on the domain size. The number of membership queries is $O(3^a)$ in the worst case (with $a := |anc^\mathbf{V}_\model(Y)|$) and reduces to a polynomial bound under bounded intervention size; the full three-parameterisation analysis is deferred to Appendix~\ref{app:complexity}. The remainder of this section describes pruning strategies that substantially reduce the practical cost without affecting soundness or completeness.

\begin{remark}[Phase~2 pruning strategies]\label{rem:pruning}
    Three strategies reduce the cost of Phase~2 without affecting soundness or completeness:
    \textbf{(i)~size-first traversal} processes smaller nodes first, letting confirmed causes prune supersets early;
    \textbf{(ii)~node-level pre-check} tests existence of $\vec{o} \in \hat{\mathcal{O}}_{\model,do(\vec{S})}(\hat{\mathcal{Y}})$ with $\vec{o}_\mathbf{U} = \vec{u}$ before enumerating the $2^{|\vec{S}|}-1$ subsets of $\vec{S}$, skipping the node on failure;
    \textbf{(iii)~superset pruning} skips any $\vec{X} \subseteq \vec{S}$ for which a confirmed $\vec{X}' \subsetneq \vec{X}$ already exists.
    The completeness proof of Theorem~\ref{thm:witness-suff} only needs nodes satisfying the condition of Proposition~\ref{prop:ancestor-reachability}, so nodes violating it never have to be tested.
    Justifications, the implication that~(iii) makes Phase~3 redundant when combined with~(i), and continuous-domain adaptations (interval, LP, and bisection-based pre-checks) are given in Appendix~\ref{app:pruning}.
\end{remark}

Appendix~\ref{app:example-forest-fire} works through the forest-fire scenario,
contrasting overdetermination ($F_Y = L \vee A$, unique joint cause) with
conjunctive causation ($F_Y = L \wedge A$, two singleton but-for causes).
Appendix~\ref{app:example-billy-suzy} (Figure~\ref{fig:billy-suzy1}) covers the
Billy-and-Suzy preemption example, illustrating the role of non-empty contingency
sets and the derivation of responsibility scores.

\subsection{Baseline Comparisons}\label{sec:experiments}

\begin{figure}[t]
    \centering
    \includegraphics[width=0.99\linewidth]{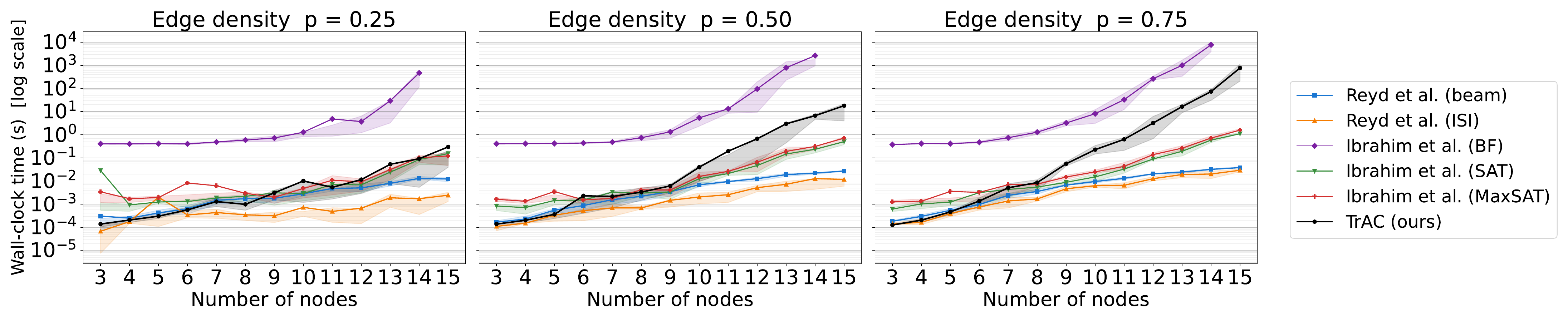}
    \caption{\textbf{Runtime comparisons to adjacent baselines.}
    Mean runtime (lines) with $[25,75]$-IQR bands (shading) across 10 seeds for
    TrAC, \citet{Reyd2025} (beam search, ISI), \& \citet{Ibrahim2020} (HP2SAT
    BF, SAT, MaxSAT) on random Boolean SCMs.
    Each SCM is drawn from an acyclic Erd\H{o}s-R\'enyi DAG with $n \in \{3,\ldots,15\}$
    nodes \& edge density $d \in \{\frac{1}{4}, \frac{1}{2}, \frac{3}{4}\}$; structural equations are
    random AND/OR gates with independently negated inputs. (Best viewed in color.)
    }
    \label{fig:trac-compare}
\end{figure}

We benchmark TrAC on random Boolean SCMs from acyclic Erd\H{o}s-R\'enyi DAGs
($n \in \{3,\ldots,15\}$, density $d \in \{0.25,0.50,0.75\}$, 10 seeds each),
comparing against \citet{Reyd2025} (beam search, ISI) and \citet{Ibrahim2020}
(HP2SAT brute-force (BF), SAT, MaxSAT). Runtimes are shown in Fig.~\ref{fig:trac-compare}.
The comparison is not perfectly direct: TrAC is the only method that provably
enumerates \emph{all} actual causes, while the baselines target a single cause
without completeness guarantees.
On these benchmarks, the combined pruning of Remark~\ref{rem:pruning}(ii+iii) removes on average over $97\%$
of witness-map nodes at $n \geq 10$ for $d \geq 0.5$ and $73$--$94\%$ for $d = 0.25$ (Fig.~\ref{fig:app-pruning}); the rate is instance-dependent and leaves the $O(3^a)$ worst case of Section~\ref{sec:algorithm} unchanged.
As a result, TrAC is typically more than an order of magnitude faster than the brute-force search, while
single-cause methods (SAT, MaxSAT, beam, ISI) are faster per query, as expected
given they solve a strictly simpler task without completeness guarantees.
Further plots and discussion are in Appendix~\ref{app:experiments}. All code is
available at \url{https://github.com/MoritzWillig/tracingActualCauses}.

\section{Conclusion}
\label{sec:conclusion}

We presented a constructive procedure for enumerating all actual causes of an event in an acyclic SCM.
The key insight is a change of direction: rather than searching forward over alternative values, we propagate the counterfactual target backward through the
structural equations.
This yields \emph{witness targets} (Definitions~\ref{def:parent-targets}--\ref{def:global-targets}, Theorem~\ref{thm:witness-target}), a criterion for counterfactual satisfaction that converts the existential search in AC2 into a single set-membership query, eliminating any dependence on domain size from the per-candidate verification cost.
An \emph{ancestor intervention grammar} then generates all structurally admissible candidate intervention sets as nodes of a finite DAG and the \emph{witness map} (Definitions~\ref{def:anc-grammar}--\ref{def:witness-map}, Propositions~\ref{prop:grammar-equiv}--\ref{prop:ancestor-reachability}), whose traversal is guaranteed to cover every actual cause (Theorem~\ref{thm:witness-suff}).
Algorithm~\ref{alg:find-causes} combines both ingredients and is proven sound and complete (Theorem~\ref{thm:sound-complete}) for the modified HP definition.

\textbf{Limitations.}
TrAC works with fully specified acyclic SCM; latent confounders can be encoded as exogenous roots, but unknown or partially specified models are out of scope.
The main computational limitation is storage: exact global targets require $O(D^{|\mathbf{O}|})$ space in the worst case, with $D$ the largest domain size, plus preimage tables of size $O(|\mathbf{V}| \cdot D^{\Delta+1})$, with $\Delta$ the largest number of endogenous parents (Appendix~\ref{app:complexity-setup}).
All 390 benchmark instances up to $n{=}15$ completed on a single 32\,GB machine (Appendix~\ref{app:experiments}), whereas our attempts at $n{=}16$ ran out of memory;
since each additional Boolean variable doubles the number of configurations, added memory buys roughly one further node.
The outer approximation of Appendix~\ref{app:constraints} needs only $O(|\mathbf{V}| \cdot D^\Delta)$ storage, but it can only screen candidates: a failed check safely rules a candidate out.
Compact exact target representations (e.g.\ BDDs, polytopes) for broader mechanism classes, full distributions $P(\mathbf{U})$ for probabilistic actual causation, and nondeterministic SCMs are the most pressing open directions (Appendix~\ref{app:roadmap}).

\begin{ack}
We thank the reviewers for their constructive and thoughtful feedback, which helped to improve this work. This work has benefited from the German Research Foundation (DFG) research grant ``Tractable Neuro-Causal Models'' (KE~1686/8-1), from the clusters of excellence ``Reasonable Artificial Intelligence'' (RAI, EXC~3057) and ``The Adaptive Mind'' (TAM, EXC~3066) under Germany's Excellence Strategy, and from the LOEWE Center DYNAMIC of the State of Hesse. It was also funded by the European Union (Grant Agreement no.~101120763, TANGO). Views and opinions expressed are, however, those of the author(s) only and do not necessarily reflect those of the European Union or the European Health and Digital Executive Agency (HaDEA). Neither the European Union nor the granting authority can be held responsible for them.
\end{ack}

\bibliography{references}

@inproceedings{Halpern2015,
  author    = {Halpern, Joseph Y.},
  title     = {A Modification of the {Halpern}-{Pearl} Definition of Causality},
  booktitle = {Proceedings of the 24th International Joint Conference on
               Artificial Intelligence},
  pages     = {3022--3033},
  year      = {2015}
}

@book{Halpern2016,
  author    = {Halpern, Joseph Y.},
  title     = {Actual Causality},
  publisher = {MIT Press},
  year      = {2016}
}

@article{HalpernPearl2005,
  title={Causes and explanations: {A} structural-model approach. {Part I}: Causes},
  author={Halpern, Joseph Y and Pearl, Judea},
  journal={The British Journal for the Philosophy of Science},
  volume={56},
  number={4},
  pages={843--887},
  year={2005},
  publisher={The University of Chicago Press}
}

@book{Pearl2009,
  title={Causality},
  author={Pearl, Judea},
  year={2009},
  publisher={Cambridge University Press}
}

@article{Eiter2002,
  title={Complexity results for structure-based causality},
  author={Eiter, Thomas and Lukasiewicz, Thomas},
  journal={Artificial Intelligence},
  volume={142},
  number={1},
  pages={53--89},
  year={2002},
  publisher={Elsevier}
}

@article{Aleksandrowicz2014,
  title={The computational complexity of structure-based causality},
  author={Aleksandrowicz, Gadi and Chockler, Hana and Halpern, Joseph Y and Ivrii, Alexander},
  journal={Journal of Artificial Intelligence Research},
  volume={58},
  pages={431--451},
  year={2017}
}

@inproceedings{Reyd2025,
  title={Searching for actual causes: Approximate algorithms with adjustable precision},
  author={Reyd, Samuel and Diaconescu, Ada and Dessalles, Jean-Louis},
  booktitle={NeurIPS 2025 Workshop on CauScien: Uncovering Causality in Science},
  year      = {2025}
}

@article{Chockler2004,
  title={Responsibility and blame: A structural-model approach},
  author={Chockler, Hana and Halpern, Joseph Y},
  journal={Journal of Artificial Intelligence Research},
  volume={22},
  pages={93--115},
  year={2004}
}

@book{Moore2009,
    author = {Moore, Michael S.},
    title = {Causation and Responsibility: {An} Essay in Law, Morals, and Metaphysics},
    publisher = {Oxford University Press},
    year = {2009},
}

@article{Cushman2008,
  title={Moral appraisals affect doing/allowing judgments},
  author={Cushman, Fiery and Knobe, Joshua and Sinnott-Armstrong, Walter},
  journal={Cognition},
  volume={108},
  number={1},
  pages={281--289},
  year={2008},
  publisher={Elsevier}
}

@article{Beckers2021,
  title={Causal sufficiency and actual causation},
  author={Beckers, Sander},
  journal={Journal of Philosophical Logic},
  volume={50},
  number={6},
  pages={1341--1374},
  year={2021},
  publisher={Springer}
}

@article{BeckersVennekens2016,
  title={A general framework for defining and extending actual causation using CP-logic},
  author={Beckers, Sander and Vennekens, Joost},
  journal={International Journal of Approximate Reasoning},
  volume={77},
  pages={105--126},
  year={2016},
  publisher={Elsevier}
}

@article{FentonGlynn2017,
  title={A Proposed Probabilistic Extension of the {Halpern} and {Pearl} Definition of `{Actual Cause}'},
  author={Fenton-Glynn, Luke},
  journal={The British Journal for the Philosophy of Science},
  volume={68},
  number={4},
  pages={1061--1124},
  year={2017}
}

@inproceedings{Beckers2025,
  title={Actual Causation and Nondeterministic Causal Models},
  author={Beckers, Sander},
  booktitle={Proceedings of the Fourth Conference on Causal Learning and Reasoning},
  series={Proceedings of Machine Learning Research},
  volume={275},
  pages={514--532},
  year={2025},
  organization={PMLR}
}

@article{BeckersChocklerHalpern2022,
  title={A causal analysis of harm},
  author={Beckers, Sander and Chockler, Hana and Halpern, Joseph Y.},
  journal={Minds and Machines},
  volume={34},
  number={3},
  pages={34},
  year={2024},
  publisher={Springer},
  doi={10.1007/s11023-024-09689-7}
}

@article{Lundberg2017,
  title={A unified approach to interpreting model predictions},
  author={Lundberg, Scott M and Lee, Su-In},
  journal={Advances in neural information processing systems},
  volume={30},
  year={2017}
}

@article{Wachter2018,
  title={Counterfactual Explanations Without Opening the Black Box: Automated Decisions and the GDPR},
  author={Wachter, Sandra and Mittelstadt, Brent and Russell, Chris},
  journal={Harvard Journal of Law \& Technology},
  volume={31},
  number={2},
  pages={841--887},
  year={2018}
}

@inproceedings{Mothilal2020,
  title={Explaining machine learning classifiers through diverse counterfactual explanations},
  author={Mothilal, Ramaravind K and Sharma, Amit and Tan, Chenhao},
  booktitle={Proceedings of the 2020 conference on fairness, accountability, and transparency},
  pages={607--617},
  year={2020}
}

@article{Karimi2022,
  title={A survey of algorithmic recourse: contrastive explanations and consequential recommendations},
  author={Karimi, Amir-Hossein and Barthe, Gilles and Sch{\"o}lkopf, Bernhard and Valera, Isabel},
  journal={ACM Computing Surveys},
  volume={55},
  number={5},
  pages={1--29},
  year={2022},
  publisher={ACM New York, NY}
}

@inproceedings{Ibrahim2020,
  title={From checking to inference: Actual causality computations as optimization problems},
  author={Ibrahim, Amjad and Pretschner, Alexander},
  booktitle={International Symposium on Automated Technology for Verification and Analysis},
  pages={343--359},
  year={2020},
  organization={Springer}
}

@inproceedings{Galhotra2021,
  title={Explaining black-box algorithms using probabilistic contrastive counterfactuals},
  author={Galhotra, Sainyam and Pradhan, Romila and Salimi, Babak},
  booktitle={Proceedings of the 2021 international conference on management of data},
  pages={577--590},
  year={2021}
}

@inproceedings{Janzing2020,
  title={Feature relevance quantification in explainable {AI}: {A} causal problem},
  author={Janzing, Dominik and Minorics, Lenon and Bl{\"o}baum, Patrick},
  booktitle={International Conference on artificial intelligence and statistics},
  pages={2907--2916},
  year={2020},
  organization={PMLR}
}

@article{Heskes2020,
  title={Causal shapley values: Exploiting causal knowledge to explain individual predictions of complex models},
  author={Heskes, Tom and Sijben, Evi and Bucur, Ioan Gabriel and Claassen, Tom},
  journal={Advances in neural information processing systems},
  volume={33},
  pages={4778--4789},
  year={2020}
}

@inproceedings{zevcevic2026causal,
  title={Causal Articulation Theory (CAT): Articulating Static and Temporal Causal Models Beyond LLM Rationalizations},
  author={Ze{\v{c}}evi{\'c}, Matej and Dhami, Devendra Singh and Kersting, Kristian},
  booktitle={Fifth Conference on Causal Learning and Reasoning},
  year={2026}
}

@inproceedings{Boutilier1996,
  title={Context-specific independence in Bayesian networks},
  author={Boutilier, Craig and Friedman, Nir and Goldszmidt, Moises and Koller, Daphne},
  booktitle={Proc. of the 13th Conf. on Uncertainty in Artificial Intelligence},
  year={1996}
}

@inproceedings{Zhang1999,
  title={On the role of context-specific independence in probabilistic inference},
  author={Zhang, Nevin Lianwen and Poole, David},
  booktitle={IJCAI},
  volume={99},
  pages={1288--1293},
  year={1999}
}

@article{Pensar2015,
  title={Labeled directed acyclic graphs: a generalization of context-specific independence in directed graphical models},
  author={Pensar, Johan and Nyman, Henrik and Koski, Timo and Corander, Jukka},
  journal={Data mining and knowledge discovery},
  volume={29},
  number={2},
  pages={503--533},
  year={2015},
  publisher={Springer}
}

@inproceedings{Budhathoki2022,
  title={Causal structure-based root cause analysis of outliers},
  author={Budhathoki, Kailash and Minorics, Lenon and Bl{\"o}baum, Patrick and Janzing, Dominik},
  booktitle={International conference on machine learning},
  pages={2357--2369},
  year={2022},
  organization={PMLR}
}

@article{Ikram2022root,
  title={Root cause analysis of failures in microservices through causal discovery},
  author={Ikram, Azam and Chakraborty, Sarthak and Mitra, Subrata and Saini, Shiv and Bagchi, Saurabh and Kocaoglu, Murat},
  journal={Advances in Neural Information Processing Systems},
  volume={35},
  pages={31158--31170},
  year={2022}
}

@article{Yang2022causal,
  title={A causal approach to detecting multivariate time-series anomalies and root causes},
  author={Yang, Wenzhuo and Zhang, Kun and Hoi, Steven CH},
  journal={arXiv preprint arXiv:2206.15033},
  year={2022}
}

@article{Kim2013root,
  title={Root cause detection in a service-oriented architecture},
  author={Kim, Myunghwan and Sumbaly, Roshan and Shah, Sam},
  journal={ACM SIGMETRICS Performance Evaluation Review},
  volume={41},
  number={1},
  pages={93--104},
  year={2013},
  publisher={ACM New York, NY, USA}
}

@inproceedings{BeckersChocklerHalpern2023quantifying,
  title={Quantifying harm},
  author={Beckers, Sander and Chockler, Hana and Halpern, Joseph Y.},
  booktitle={Proceedings of the Thirty-Second International Joint Conference on Artificial Intelligence (IJCAI)},
  pages={363--371},
  year={2023},
  doi={10.24963/ijcai.2023/41}
}

@inproceedings{ChocklerHalpern2024image,
  title={Explaining Image Classifiers},
  author={Chockler, Hana and Halpern, Joseph Y.},
  booktitle={Proceedings of the 21st International Conference on Principles of Knowledge Representation and Reasoning (KR)},
  pages={264--272},
  year={2024}
}

\newpage

\clearpage
\section*{Appendix for ``Tracing Actual Causes with Counterfactual Witness Maps''}
\vspace{4cm}
\DoToC
\clearpage

\appendix

\section{Notation}\label{app:notation}

\begin{table}[h]\vspace{-3\baselineskip}
\centering
\caption{Notation summary.}\label{tab:notation}
\small
\setlength{\tabcolsep}{4pt}
\renewcommand{\arraystretch}{0.9}
\begin{tabular}{ll}
\toprule
\textbf{Symbol} & \textbf{Meaning} \\
\midrule
\multicolumn{2}{l}{\textit{Variables and subsets}} \\
$\mathbf{V}, \mathbf{U}$ & sets of endogenous, exogenous variables \\
$\mathbf{O} = \mathbf{U} \cup \mathbf{V}$ & set of all variables (full configurations) \\
$X, Y, X_i$ & single variables \\
$\vec{X}, \vec{Z}, \vec{S}$ & arbitrary subsets of $\mathbf{V}$ ($\vec{Z}$ used as contingency, $\vec{S}$ as intervention/node) \\
$pa(X), ch(X), anc(X)$ & parents, children, ancestors of $X$ ($pa^\mathbf{V}_\model, anc^\mathbf{V}_\model$ restricted to endogenous) \\
$pa^\mathbf{V}_\model(\vec{S})$ & set form $\bigcup_{X \in \vec{S}} pa^\mathbf{V}_\model(X)$ \\
$\mathcal{P}(\cdot)$ & powerset \\
$\mathcal{G}_\model$ & causal DAG induced by $\model$ \\
\midrule
\multicolumn{2}{l}{\textit{Values and domains}} \\
$x, y, x_i$ & values of single variables \\
$\vec{x}, \vec{z}, \vec{s}$ & joint value tuples for $\vec{X}, \vec{Z}, \vec{S}$ \\
$\vec{u}$ & environment (joint value of $\mathbf{U}$) \\
$\vec{o}$ & joint outcome (joint value of $\mathbf{O}$) \\
$\vec{o}_{\vec{u},\vec{s}}$ & unique joint configuration of $(\model_{do(\vec{S}=\vec{s})}, \vec{u})$ \\
$\mathcal{X}_i$ & domain of single variable $X_i$ \\
$\vec{\mathcal{X}}, \vec{\mathcal{S}}$ & joint-value domains for $\vec{X}, \vec{S}$ (canonical); $\vec{\mathcal{X}}_{pa(X)}$ analogously for $pa(X)$ \\
$\vec{\mathcal{U}}, \vec{\mathcal{V}}, \vec{\mathcal{O}}$ & joint-value domains for $\mathbf{U}, \mathbf{V}, \mathbf{O}$ \\
$\vec{o}_\mathbf{U}, \vec{o}_{\vec{Z}}$ & projection of $\vec{o}$ onto components $\mathbf{U}$, $\vec{Z}$ \\
\midrule
\multicolumn{2}{l}{\textit{SCM, settings, intervention}} \\
$\model = \langle \mathbf{U}, \mathbf{V}, \mathcal{F}, \mathbb{P}_{\mathbf{U}} \rangle$ & structural causal model \\
$\mathcal{F} = \{F_X : X \in \mathbf{V}\}$ & structural equations \\
$F_X : \vec{\mathcal{X}}_{pa(X)} \to \mathcal{X}_X$ & structural equation for $X$ \\
$F_X^{-1}(x)$ & preimage of value $x$ under $F_X$ \\
$\world : \mathbf{V} \to \bigcup_{X_i \in \mathbf{V}} \mathcal{X}_i$ & world (assignment of endogenous values) \\
$\world_{\vec{u}},\world_{\vec{u}}(\vec{Z})$ & actual world induced by $\vec{u}$; actual values of $\vec{Z}$ (restriction of $\world_{\vec{u}}$ to $\vec{Z}$) \\
$\world^*_{\vec{u}}$ & counterfactual world (after intervention) \\
$(\model, \vec{u})$ & setting \\
$\model_{do(\vec{X}=\vec{x})}$ & intervened SCM \\
$\model, \vec{u} \models \varphi$ & event $\varphi$ holds in $(\model, \vec{u})$ \\
$\model, \vec{u} \models [\vec{X}=\vec{x}]\varphi$ & $\varphi$ holds in $(\model_{do(\vec{X}=\vec{x})}, \vec{u})$ \\
\midrule
\multicolumn{2}{l}{\textit{Actual causation}} \\
$(Y \in \mathcal{Y}^*)$, $(Y \in \hat{\mathcal{Y}})$ & target event, counterfactual target ($\hat{\mathcal{Y}} := \mathcal{Y} \setminus \mathcal{Y}^*$) \\
$(\vec{X} = \vec{x})$ & candidate (actual) cause \\
$\vec{Z}$ & contingency set ($\vec{Z} \subseteq \mathbf{V} \setminus \vec{X}$) \\
$(\vec{Z}, \vec{X}, \vec{x}', \mathcal{Y}^*)$ & counterfactual witness tuple \\
AC1, AC2, AC2$^*$, AC3 & modified HP conditions for actual causation \\
\midrule
\multicolumn{2}{l}{\textit{Witness targets (Section~\ref{sec:witness-targets})}} \\
$\hat{\mathcal{O}}^{pa}_\model(X, \hat{\mathcal{O}})$ & parent targets (Definition~\ref{def:parent-targets}) \\
$\hat{\mathcal{O}}^{\downarrow}_{\model,do(\vec{S})}(X, \hat{\mathcal{Y}})$ & descendant targets (Definition~\ref{def:desc-targets}) \\
$\hat{\mathcal{O}}_{\model,do(\vec{S})}(\hat{\mathcal{Y}})$ & global targets (Definition~\ref{def:global-targets}) \\
\midrule
\multicolumn{2}{l}{\textit{Grammar and witness map (Section~\ref{sec:witness-maps})}} \\
$\Sigma_\model(Y)$ & alphabet $anc^\mathbf{V}_\model(Y) \cup \{Y\}$ of the grammar \\
$G_\model(Y), \mathcal{L}(G_\model(Y))$ & ancestor intervention grammar, language (Definition~\ref{def:anc-grammar}) \\
$(P_{\mathrm{keep}}), (P_{\mathrm{unfold}})$ & keep / unfold productions \\
$\vec{S}_{\mathrm{unfold}}, \vec{S}_{\mathrm{keep}}$ & unfold / keep partition of a node in a derivation step \\
$\Rightarrow_G, \Rightarrow_G^{+}$ & one-step / multi-step derivation \\
$\mathcal{A}_\model(Y)$ & ancestral map (Definition~\ref{def:anc-map}) \\
$ch_{\mathcal{A}}(\vec{S})$ & children of node $\vec{S}$ in the ancestral map \\
$\mathcal{W}_\model(Y \in \hat{\mathcal{Y}})$ & witness map (Definition~\ref{def:witness-map}) \\
$\mathcal{C}$ & set of actual causes returned by Algorithm~\ref{alg:find-causes} \\
\bottomrule
\end{tabular}
\end{table}

\section{Extended Proofs}\label{app:proofs}

\subsection*{Proof of Theorem~\ref{thm:witness-target} (Witness Target Membership)}

\begin{proof}
    We proceed by induction on the reverse topological order of the DAG, traversing from the target $Y$ upward to the roots of $\model_{do(\vec{S})}$.

    \noindent\emph{($\Rightarrow$)} Suppose $\vec{o}_{\vec{u},\vec{s}} \in \hat{\mathcal{O}}_{\model,do(\vec{S})}(\hat{\mathcal{Y}})$. By Definition~\ref{def:global-targets}, $\vec{o}_{\vec{u},\vec{s}} \in \hat{\mathcal{O}}^{\downarrow}_{\model,do(\vec{S})}(R,\hat{\mathcal{Y}})$ for every root $R \in \mathbf{U} \cup \vec{S}$. Since every endogenous variable has a parent, following parents upward from $Y$ through non-intervened variables reaches such a root along a directed path $R \to X_1 \to \dots \to X_k = Y$; since $\hat{\mathcal{O}}^{pa}(X_j, \hat{\mathcal{O}}) \subseteq \hat{\mathcal{O}}$, the recursion of Definition~\ref{def:desc-targets} gives $\hat{\mathcal{O}}^{\downarrow}(R) \subseteq \hat{\mathcal{O}}^{\downarrow}(X_1) \subseteq \dots \subseteq \hat{\mathcal{O}}^{\downarrow}(Y)$. Hence the base case of Definition~\ref{def:desc-targets} at $X = Y$ requires $(\vec{o}_{\vec{u},\vec{s}})_Y \in \hat{\mathcal{Y}}$, i.e.\ $\world^*_{\vec{u}}(Y) \in \hat{\mathcal{Y}}$, which gives $\model_{do(\vec{S}=\vec{s})}, \vec{u} \models (Y \in \hat{\mathcal{Y}})$.

    \noindent\emph{($\Leftarrow$)} Suppose $\model_{do(\vec{S}=\vec{s})}, \vec{u} \models (Y \in \hat{\mathcal{Y}})$, i.e.\ $\world^*_{\vec{u}}(Y) \in \hat{\mathcal{Y}}$. Since $\model$ is acyclic, $\vec{o}_{\vec{u},\vec{s}}$ is the unique joint configuration consistent with $\vec{u}$ and the structural equations of $\model_{do(\vec{S}=\vec{s})}$. We show $\vec{o}_{\vec{u},\vec{s}} \in \hat{\mathcal{O}}^{\downarrow}_{\model,do(\vec{S})}(X,\hat{\mathcal{Y}})$ for every $X \in \mathbf{V} \cup \mathbf{U}$ by induction on reverse topological order (from $Y$ upward):
    \textbf{Base case} $X = Y$: By the first branch of Definition~\ref{def:desc-targets}, $\hat{\mathcal{O}}^{\downarrow}(Y) = \{\vec{o} \in \vec{\mathcal{O}} \mid \vec{o}_Y \in \hat{\mathcal{Y}}\}$, and $(\vec{o}_{\vec{u},\vec{s}})_Y = \world^*_{\vec{u}}(Y) \in \hat{\mathcal{Y}}$ by assumption.
    \textbf{Inductive step}: Assume the claim holds for all children $Z \in ch(X) \setminus \vec{S}$. By definition, $\hat{\mathcal{O}}^{\downarrow}(X) = \bigcap_{Z \in ch(X) \setminus \vec{S}} \hat{\mathcal{O}}^{pa}(Z,\,\hat{\mathcal{O}}^{\downarrow}(Z))$. For each such $Z$, the inductive hypothesis gives $\vec{o}_{\vec{u},\vec{s}} \in \hat{\mathcal{O}}^{\downarrow}(Z)$. Furthermore, the structural equation $F_Z$ evaluated at the actual parent configuration $(\vec{o}_{\vec{u},\vec{s}})_{pa(Z)}$ produces $(\vec{o}_{\vec{u},\vec{s}})_Z$ by construction, so $(\vec{o}_{\vec{u},\vec{s}})_{pa(Z)} \in F_Z^{-1}((\vec{o}_{\vec{u},\vec{s}})_Z)$ and hence $\vec{o}_{\vec{u},\vec{s}} \in \hat{\mathcal{O}}^{pa}(Z,\,\hat{\mathcal{O}}^{\downarrow}(Z))$ (Definition~\ref{def:parent-targets}). The intersection over all non-intervened children yields $\vec{o}_{\vec{u},\vec{s}} \in \hat{\mathcal{O}}^{\downarrow}(X)$.
    \textbf{Leaves} $ch(X) = \emptyset$: The second branch of Definition~\ref{def:desc-targets} sets $\hat{\mathcal{O}}^{\downarrow}(X) = \vec{\mathcal{O}}$, which trivially contains $\vec{o}_{\vec{u},\vec{s}}$. Taking the intersection over all roots $R \in \mathbf{U} \cup \vec{S}$ (Definition~\ref{def:global-targets}) yields $\vec{o}_{\vec{u},\vec{s}} \in \hat{\mathcal{O}}_{\model,do(\vec{S})}(\hat{\mathcal{Y}})$.
\end{proof}

\subsection*{Proof of Proposition~\ref{prop:grammar-equiv} (Grammar--Graph Equivalence)}

\begin{proof}
    We show that the closed-form expression of Proposition~\ref{prop:grammar-equiv} characterises exactly the one-step derivates of $\vec{S}$ under $G_\model(Y)$. ($\supseteq$) Let $\vec{S}'' \in \mathcal{P}(pa^\mathbf{V}_\model(\vec{S}') \cup (\vec{S} \setminus \vec{S}')) \setminus \{\emptyset\}$ for some non-empty $\vec{S}' \subseteq \vec{S}$. Set $\vec{S}_{\mathrm{unfold}} := \vec{S}'$ and $\vec{S}_{\mathrm{keep}} := \vec{S} \setminus \vec{S}'$, so that the candidate set in \eqref{eq:deriv} is $pa^\mathbf{V}_\model(\vec{S}') \cup (\vec{S} \setminus \vec{S}')$ and $\vec{S}''$ is a non-empty subset of this candidate, hence $\vec{S} \Rightarrow_G \vec{S}''$, i.e.\ $\vec{S}'' \in ch_{\mathcal{A}}(\vec{S})$. ($\subseteq$) Let $\vec{S}'' \in ch_{\mathcal{A}}(\vec{S})$, i.e.\ $\vec{S} \Rightarrow_G \vec{S}''$ by Definition~\ref{def:anc-map}, via partition $\vec{S}_{\mathrm{unfold}} \;\dot{\cup}\; \vec{S}_{\mathrm{keep}}$ with $\vec{S}_{\mathrm{unfold}} \neq \emptyset$. Setting $\vec{S}' := \vec{S}_{\mathrm{unfold}}$ in the closed-form union implies $\vec{S}_{\mathrm{keep}} = \vec{S} \setminus \vec{S}'$ and $\vec{S}''$ being a non-empty subset of $pa^\mathbf{V}_\model(\vec{S}') \cup (\vec{S} \setminus \vec{S}')$. Hence, $\vec{S}''$ lies in the right-hand side.
\end{proof}

\subsection*{Proof of Lemma~\ref{lem:ac-witness} (AC-Witness Identification)}

\begin{proof}
    \textbf{(AC1):} By assumption, $\model, \vec{u} \models (Y \in \mathcal{Y}^*)$. Moreover, $\model, \vec{u} \models (\vec{X} = \world_{\vec{u}}(\vec{X}))$ holds trivially, since $\world_{\vec{u}}$ is the actual world induced by $\vec{u}$. Hence $\model, \vec{u} \models (\vec{X} = \world_{\vec{u}}(\vec{X})) \land (Y \in \mathcal{Y}^*)$, which is $(AC1)$.
    \textbf{(AC2)/(AC2$^*$):} By Corollary~\ref{cor:witness-existence}, the existence of $\vec{o} \in \hat{\mathcal{O}}_{\model,do(\vec{X},\vec{Z})}(\hat{\mathcal{Y}})$ with $\vec{o}_\mathbf{U} = \vec{u}$ and $\vec{o}_{\vec{Z}} = \world_{\vec{u}}(\vec{Z})$ gives the existence of $\vec{x}' \in \vec{\mathcal{X}}$ such that $\model, \vec{u} \models [\vec{X} = \vec{x}',\vec{Z} = \world_{\vec{u}}(\vec{Z})](Y \in \hat{\mathcal{Y}})$. It remains to verify $\vec{x}' \neq \world_{\vec{u}}(\vec{X})$. If $\vec{x}' = \world_{\vec{u}}(\vec{X})$, then $\model_{do(\vec{X}=\vec{x}', \vec{Z}=\world_{\vec{u}}(\vec{Z}))}, \vec{u}$ agrees with the actual setting $(\model, \vec{u})$ by the \emph{consistency} of structural equations, so $Y$ takes its actual value $\world_{\vec{u}}(Y) \in \mathcal{Y}^*$, contradicting $Y \in \hat{\mathcal{Y}} = \mathcal{Y} \setminus \mathcal{Y}^*$. Therefore $\vec{x}' \neq \world_{\vec{u}}(\vec{X})$, and $(\vec{Z}, \vec{X}, \vec{x}', \mathcal{Y}^*)$ constitutes a valid counterfactual witness for (AC2). For (AC2$^*$), the same argument applies with $\vec{Z} = \emptyset$. In both cases, $(\vec{X} = \world_{\vec{u}}(\vec{X}))$ is a but-for or actual cause, respectively, upon applying the minimality filter $(AC3)$.
\end{proof}

\subsection{Counterexample to the Necessity of Proposition~\ref{prop:ancestor-reachability}}\label{app:prop4-counterex}

The path-disjointness condition of Proposition~\ref{prop:ancestor-reachability} is sufficient but not necessary for $\vec{S} \in \mathcal{A}_\model(Y)$. Consider the DAG with edges $A \to B$, $B \to X_1$, $B \to X_2$, $X_1 \to Y$, $X_2 \to Y$. The ancestor set $\vec{S} = \{A, B\}$ is reachable in $\mathcal{A}_\model(Y)$ via the derivation
\[
\{Y\} \;\Rightarrow\; \{X_1, X_2\} \;\Rightarrow\; \{B, X_2\} \;\Rightarrow\; \{A, B\},
\]
where the diamond structure allows $B$ to re-enter the set after being unfolded (the last step unfolds both $B$ and $X_2$, with $pa(B) \cup pa(X_2) = \{A\} \cup \{B\} = \{A, B\}$). Yet every directed path from $A$ to $Y$ passes through $B$, so the path-disjointness condition fails on $A$ while $\{A, B\} \in \mathcal{A}_\model(Y)$. Reachability of an ancestor set is therefore generated by a non-local sequence of grammar steps that can re-introduce intermediate variables; it is not captured by any local property of cause-to-effect path structure.

\section{Examples}\label{app:examples}
\subsection{Forest-Fire Example}\label{app:example-forest-fire}

We use the forest-fire scenario to contrast two qualitatively different causal
structures that share the same DAG but differ in the structural equation for $Y$.
Both cases use the DAG $L \to Y \leftarrow A$ with binary variables
$\mathcal{X}_i = \{0,1\}$, independent exogenous roots $U_L, U_A$, and equations
$F_L = U_L$, $F_A = U_A$. The setting $\vec{u}$ has $U_L = U_A = 1$, so
$\world_{\vec{u}} = (L, A, Y) = (1, 1, 1)$. We explain $\mathcal{Y}^* = \{1\}$,
hence $\hat{\mathcal{Y}} = \{0\}$.

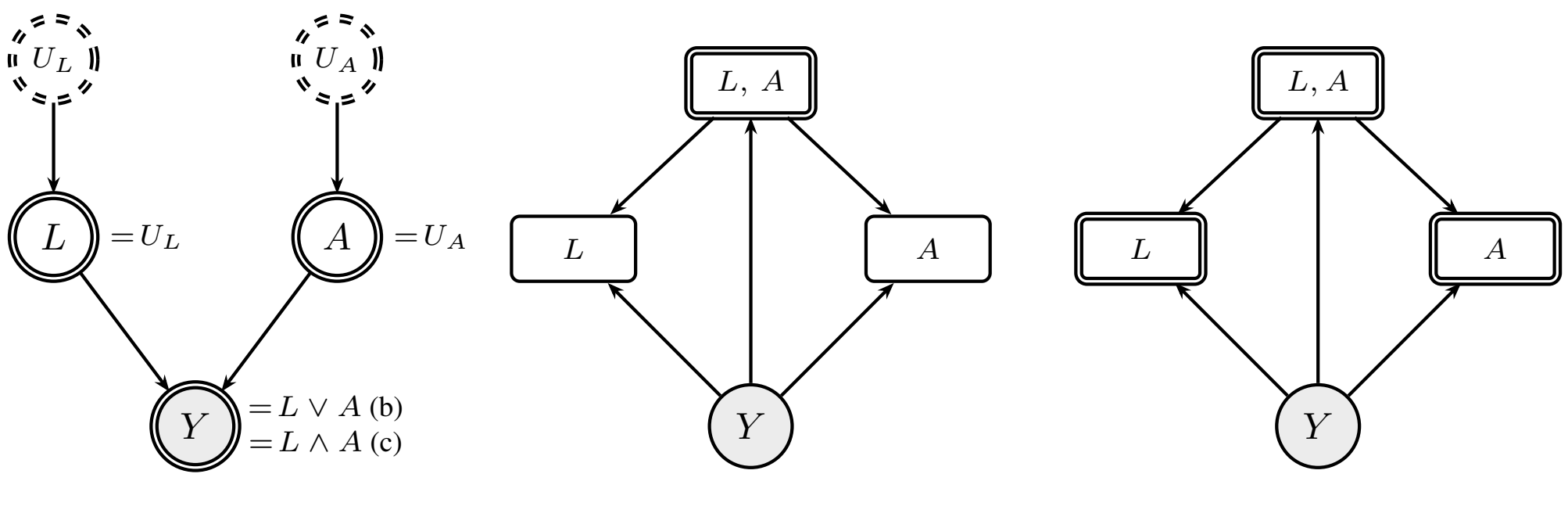
\begin{figure}[h]
\centering
\begin{tikzpicture}[
    dag node/.style    = {circle, draw, thick, minimum size=0.70cm,
                          inner sep=1pt, font=\small},
    exo node/.style    = {dag node, dashed},
    lattice node/.style= {rectangle, draw, thick, rounded corners=2pt,
                          minimum width=1.05cm, minimum height=0.55cm,
                          font=\scriptsize, inner sep=2pt},
    lbl/.style         = {draw=none, fill=none, font=\scriptsize,
                          inner sep=1pt},
    arr/.style         = {-{Stealth[length=4pt]}, thick},
]

\begin{scope}[xshift=0cm]
  \node[font=\footnotesize\bfseries] at (0, 0) {(a)};
  \node[dag node, fill=gray!15, double]  (Y)  at ( 0.0, 1.0) {$Y$};
  \node[dag node, double]                (L)  at (-1.2, 2.6) {$L$};
  \node[dag node, double]                (A)  at ( 1.2, 2.6) {$A$};
  \node[exo node, double]                (UL) at (-1.2, 4.1) {\scriptsize$U_L$};
  \node[exo node, double]                (UA) at ( 1.2, 4.1) {\scriptsize$U_A$};
  \draw[arr] (UL)--(L);  \draw[arr] (UA)--(A);
  \draw[arr] (L)--(Y);   \draw[arr] (A)--(Y);
  \node[lbl, right=0.07cm of L]  {\scriptsize$=\!U_L$};
  \node[lbl, right=0.07cm of A]  {\scriptsize$=\!U_A$};
  \node[lbl] (l1) at ( 1.1, 1.15) {\scriptsize$=\!L \vee A$ (b)};
  \node[lbl] (l2) at ( 1.1, .85) {\scriptsize$=\!L \wedge A$ (c)};
\end{scope}

\begin{scope}[xshift=4.7cm]
  \node[font=\footnotesize\bfseries] at (0, 0) {(b) $F_Y = L \vee A$};
  \node[dag node, fill=gray!15] (bY)  at ( 0.0, 1.0) {$Y$};
  \node[lattice node]           (bL)  at (-1.5, 2.5) {$L$};
  \node[lattice node]           (bA)  at ( 1.5, 2.5) {$A$};
  \node[lattice node, double]   (bLA) at ( 0.0, 3.9) {$L,\, A$};
  \draw[arr] (bY)--(bL);  \draw[arr] (bY)--(bA);  \draw[arr] (bY)--(bLA);
  \draw[arr] (bLA)--(bL); \draw[arr] (bLA)--(bA);
\end{scope}

\begin{scope}[xshift=9.5cm]
  \node[font=\footnotesize\bfseries] at (0, 0) {(c) $F_Y = L \wedge A$};
  \node[dag node, fill=gray!15]         (cY)  at ( 0.0, 1.0) {$Y$};
  \node[lattice node, double]           (cL)  at (-1.5, 2.5) {$L$};
  \node[lattice node, double]           (cA)  at ( 1.5, 2.5) {$A$};
  \node[lattice node, double]
                                        (cLA) at ( 0.0, 3.9) {$L,A$};
  \draw[arr] (cY)--(cL);  \draw[arr] (cY)--(cA);  \draw[arr] (cY)--(cLA);
  \draw[arr] (cLA)--(cL); \draw[arr] (cLA)--(cA);
\end{scope}

\end{tikzpicture}
\caption{(a)~Shared DAG for the forest-fire scenario. All variables in the actual
world are double-bordered as they are true $(U_L,U_A,L,A,Y)=(1,1,1,1,1)$.
(b)~Witness map for Case~I ($F_Y = L\vee A$, symmetric overdetermination):
singleton nodes $\{L\}$ and $\{A\}$ fail the node-level pre-check; only the joint node $\{L,A\}$ passes.
(c)~Witness map for Case~II ($F_Y = L\wedge A$, conjunctive causation):
all non-root nodes pass but the joint node is non-minimal.}
\label{fig:forest-fire-app}
\end{figure}

\begin{example}[Forest fire: overdetermination and joint causes]
\label{ex:forest-fire-app}
The DAG, actual world, and witness maps for both cases are shown in
Figure~\ref{fig:forest-fire-app}.

\textbf{Case I: Overdetermination} ($F_Y = L \lor A$).
Both a lightning strike ($L = 1$) and an arsonist ($A = 1$) simultaneously ignite
fires, each alone sufficient to burn down the forest.
Backward propagation gives
$\hat{\mathcal{O}}_{do(\emptyset)}(\{0\}) = \{(0,0,0)\}$;
the actual world $(1,1,1)$ is absent.

\emph{No singleton but-for causes.}
At node $\vec{S} = \{L\}$: $do(L=0)$ gives $F_Y = 0 \lor 1 = 1$,
so $\world^*_{\vec{u}} = (0,1,1) \notin \hat{\mathcal{Y}}$: $\times$.
The node-level pre-check (Remark~\ref{rem:pruning}(ii)) fails
because every $\vec{o} \in \hat{\mathcal{O}}_{do(\{L\})}(\{0\})$
requires $U_A = 0$ (since $A = U_A$ keeps $F_Y = 0 \lor 1 = 1$ unless $A = 0$),
whereas $\vec{u}$ has $U_A = 1$.
By symmetry, $\vec{S} = \{A\}$ fails identically.

\emph{One joint actual cause.}
At the joint node $\vec{S} = \{L, A\}$, both $L$ and $A$ are intervened, so $U_L$ and
$U_A$ are disconnected from $Y$: the global target leaves the environment unconstrained and,
by propagation from the intervened roots $L$ and $A$, contains exactly the configurations with $L = A = Y = 0$.
The node-level pre-check therefore passes for any $\vec{u}$, whereas the partition
$\vec{X} = \{L\}$, $\vec{Z} = \{A\}$ fails, since it would require $A = \world_{\vec{u}}(A) = 1$.
Partition $\vec{X} = \{L, A\}$, $\vec{Z} = \emptyset$: $do(L=0, A=0) \Rightarrow
Y = 0 \lor 0 = 0 \in \hat{\mathcal{Y}}$: \checkmark.
No proper subset of $\{L,A\}$ satisfies AC2, so minimality
holds: \emph{$(L{=}1, A{=}1)$ is the unique joint actual cause of $(Y{=}1)$}.
Following \citet{Halpern2016} (Definition~6.1.6), responsibility under the modified HP definition is $\rho = 1/(|\vec{Z}|+|\vec{X}|)$
for a cause $\vec{X}$ with contingency $\vec{Z}$; here $|\vec{X}|=2$, $\vec{Z}=\emptyset$, giving
$\rho = 1/2$ per conjunct.

\textbf{Case II: Conjunctive causation} ($F_Y = L \land A$).
Both ignition sources are now necessary.
$\hat{\mathcal{O}}_{do(\emptyset)}(\{0\}) = \{(0,0,0),(1,0,0),(0,1,0)\}$;
the actual world $(1,1,1)$ is again absent.

\emph{Two singleton but-for causes.}
For $\vec{S} = \{L\}$: $do(L=0)$ gives $F_Y = 0 \land 1 = 0 \in \hat{\mathcal{Y}}$:
\checkmark.
Symmetrically, $\vec{S} = \{A\}$: \checkmark.
Both singletons are minimal, so
$(L{=}1)$ and $(A{=}1)$ are the \emph{two distinct but-for causes} of $(Y{=}1)$,
each with $\vec{Z} = \emptyset$ and responsibility $\rho = 1$.

\emph{Joint node pruned by minimality.}
The node $\vec{S} = \{L, A\}$ also passes the pre-check,
and $do(L=0, A=0) \Rightarrow Y = 0$: \checkmark.
However, $\{L\} \subsetneq \{L,A\}$ and $\{L\} \in \mathcal{C}$, so Phase~3
removes $\{L,A\}$ as non-minimal.

\emph{Summary.}
In Case~I, the unique cause is the joint conjunction $(L{=}1, A{=}1)$.
In Case~II, each variable is individually necessary, yielding two singleton
but-for causes.
In both cases the result is read off the witness map without any forward
simulation over alternative values.
\end{example}

\subsection{Billy-and-Suzy Example}\label{app:example-billy-suzy}

\begin{example}[Billy and Suzy: preemption and contingency]
\label{ex:billy-suzy}
The Billy-and-Suzy scenario is the canonical illustration of \emph{preemption}:
an actual cause whose counterfactual dependence is masked by a backup process,
so that a non-empty contingency set is required to reveal it.
Suzy ($ST = 1$) and Billy ($BT = 1$) both throw rocks at a bottle.
Suzy's rock arrives first: her hit ($SH = 1$) shatters the bottle ($BS = 1$),
preempting Billy's rock from ever connecting ($BH = 0$).

\noindent
The DAG and witness map are shown in Figure~\ref{fig:billy-suzy1}.
The SCM has binary variables $\mathcal{X}_i = \{0,1\}$, independent exogenous
roots $U_{ST}, U_{BT}$, and structural equations
\[
  F_{SH} = ST, \;
  F_{BH} = BT \land \lnot SH, \;
  F_{BS} = SH \lor BH.
\]
In the setting $\vec{u}$ with $U_{ST} = U_{BT} = 1$ the actual world is
\[
  \world_{\vec{u}} = (ST, BT, SH, BH, BS) = (1, 1, 1, 0, 1).
\]
We explain $\mathcal{Y}^* = \{1\}$, so $\hat{\mathcal{Y}} = \{0\}$.

\textbf{Observational global targets.}
Backward propagation of $\hat{\mathcal{Y}} = \{0\}$ through $F_{BS} = SH \lor BH$
requires $SH = BH = 0$.
The set $\hat{\mathcal{O}}_{do(\emptyset)}(\{0\})$ therefore contains
exactly those worlds with $SH = BH = 0$; since $\world_{\vec{u}}$ has $SH = 1$, the
actual world is absent, confirming that AC1 holds and the search is non-trivial.

\textbf{No singleton but-for causes.}
All four singleton but-for tests fail:
$do(ST\!=\!0)$ gives $BH = 1 \land \lnot 0 = 1$, so $BS = 0 \lor 1 = 1$: $\times$.
$do(BT\!=\!0)$ gives $BH = 0 \land \lnot 1 = 0$, so $BS = 1 \lor 0 = 1$: $\times$.
$do(SH\!=\!0)$ gives $BH = 1 \land \lnot 0 = 1$, so $BS = 0 \lor 1 = 1$: $\times$.
$do(BH\!=\!0)$ leaves $BS = SH \lor 0 = 1$: $\times$.
Suzy's preemption of Billy's path, and Billy's backup of Suzy's, each mask the
singleton counterfactual dependencies.

\textbf{Four joint but-for causes.}
Moving to pairs in $\mathcal{A}_\model(BS)$, the node-level pre-check
(Remark~\ref{rem:pruning}(ii)) passes for exactly the four double-bordered nodes
visible in Figure~\ref{fig:billy-suzy1}(b).
Setting both members to 0 in each case witnesses $\hat{\mathcal{Y}} = \{0\}$:
\begin{itemize}
  \item $do(ST\!=\!0, BT\!=\!0) \Rightarrow SH=0,\, BH=0,\, BS=0$. \checkmark
  \item $do(ST\!=\!0, BH\!=\!0) \Rightarrow SH=0,\, BS=0$. \checkmark
  \item $do(SH\!=\!0, BT\!=\!0) \Rightarrow BH=0,\, BS=0$. \checkmark
  \item $do(SH\!=\!0, BH\!=\!0) \Rightarrow BS=0$. \checkmark
\end{itemize}
No pair is a strict subset of another, so all four are minimal: they are the
\emph{four joint but-for causes} of $(BS = 1)$.

\textbf{Two actual causes via non-empty contingency.}
The two red-highlighted nodes in Figure~\ref{fig:billy-suzy1}(b), which are $\{ST,BH\}$
and $\{SH,BH\}$, additionally admit singleton-cause partitions with contingency
$\vec{Z} = \{BH\}$, where $\world_{\vec{u}}(BH) = 0$.

\emph{Actual cause test for $ST = 1$}, node $\vec{S} = \{ST, BH\}$, partition $\vec{X} = \{ST\}$,
$\vec{Z} = \{BH\}$:
In $\model_{do(ST=0,\; BH=0)}$: $SH = 0$, $BH = 0$ (fixed by contingency),
$BS = 0 \lor 0 = 0 \in \hat{\mathcal{Y}}$.
Hence there exists $\vec{o} \in \hat{\mathcal{O}}_{do(\{ST,BH\})}(\{0\})$ with $\vec{o}_\mathbf{U} = \vec{u}$ and $\vec{o}_{BH} = \world_{\vec{u}}(BH)$:
\checkmark.
Since $\{ST\}$ is minimal, $(ST = 1)$ is an \emph{actual cause} of $(BS = 1)$
with responsibility $\rho = 1/(|\vec{Z}|+|\vec{X}|) = 1/2$.

\emph{Actual cause test for $SH = 1$}, node $\vec{S} = \{SH, BH\}$, partition $\vec{X} = \{SH\}$,
$\vec{Z} = \{BH\}$:
In $\model_{do(SH=0,\; BH=0)}$: $BH = 0$ (fixed by contingency),
$BS = 0 \lor 0 = 0 \in \hat{\mathcal{Y}}$.
Hence there exists $\vec{o} \in \hat{\mathcal{O}}_{do(\{SH,BH\})}(\{0\})$ with $\vec{o}_\mathbf{U} = \vec{u}$ and $\vec{o}_{BH} = \world_{\vec{u}}(BH)$:
\checkmark.
Since $\{SH\}$ is minimal, $(SH = 1)$ is an \emph{actual cause} of $(BS = 1)$
with responsibility $\rho = 1/(|\vec{Z}|+|\vec{X}|) = 1/2$.

Intuitively, Suzy's throw ($ST=1$) and her hit ($SH=1$) are both actual causes:
each alone, with Billy's miss held fixed at its actual value, suffices to counterfactually explain the broken bottle.
Their equal responsibilities reflect that the contingency $\{BH\}$ is the same
for both, a consequence of the preemption structure $F_{BH} = BT \land \lnot SH$.

\textbf{Why $(BT = 1)$ is not an actual cause.}
For every node $\vec{S} \in \mathcal{A}_\model(BS)$ containing $BT$ and every subset $\vec{X}$ including $BT$ with contingency $\vec{Z} = \vec{S} \setminus \vec{X}$, holding $\vec{Z}$ at actual values and setting $do(BT=0)$ leaves Suzy's path $ST \to SH \to BS$ intact.
Since $SH = 1$ in the actual world, $BS = SH \lor BH = 1$ regardless of $BT$'s
value, and no contingency over non-Suzy variables can block this route without
altering actual values.
No membership test passes for any node containing $BT$: $(BT = 1)$ is
\emph{not} an actual cause, correctly reflecting that Billy's rock never
reached the bottle.

\textbf{Summary.}
Algorithm~\ref{alg:find-causes} returns two actual causes,
$(ST = 1)$ and $(SH = 1)$, each with contingency $\vec{Z} = \{BH\}$ and
responsibility $\rho = 1/2$.
The four joint but-for causes and the two actual causes are read off directly
from Figure~\ref{fig:billy-suzy1}(b) without any forward simulation over
alternative values: the contingency set $\vec{Z} = \{BH\}$ is \emph{discovered
automatically} by the witness map traversal.
\end{example}

\section{Constraint-Based Outer Approximation}\label{app:constraints}

As an efficient alternative to computing the exact global targets, we can track \emph{per-variable marginal subsets} $\hat{\mathcal{O}} = \{\hat{\mathcal{X}}_i \mid X_i \in \mathbf{U} \cup \mathbf{V},\, \hat{\mathcal{X}}_i \subseteq \mathcal{X}_i\}$, which is a collection of individual domain subsets rather than joint configurations. This yields an \emph{outer approximation} of the global targets that is cheaper to compute and store, at the cost of possible false positives.
We re-use the operator names $\hat{\mathcal{O}}^{pa}_\model$ and $\hat{\mathcal{O}}^{\downarrow}_{\model,do(\vec{S})}$ from Definitions~\ref{def:parent-targets}--\ref{def:desc-targets} for their constraint analogues throughout this appendix; the marginal nature of the codomain is implicit in the input being a constraint collection rather than a joint set.

\begin{definition}[parent constraints]\label{def:parent-constraints}
    For SCM $\model$, variable $X \in \mathbf{V}$, and constraint collection $\hat{\mathcal{O}}$, the \textbf{parent constraints} $\hat{\mathcal{O}}^{pa}_{\model}(X, \hat{\mathcal{O}})$ are defined by marginal preimage projection:
    \begin{align}
        \hat{\mathcal{O}}^{pa}_{\model}(X, \hat{\mathcal{O}}) :=
        \Bigl(\bigl\{\bigl(F_X^{-1}(\hat{\mathcal{O}}_X)\bigr)_i \mid X_i \in pa(X)\bigr\}
        \cup \bigl\{\mathcal{X}_j \mid X_j \in (\mathbf{U} \cup \mathbf{V}) \setminus (X \cup pa(X))\bigr\}\Bigr)
        \sqcap \hat{\mathcal{O}},
    \end{align}
    where $(F_X^{-1}(\hat{\mathcal{O}}_X))_i := \{x_i \in \mathcal{X}_i \mid \exists\, \vec{o}_{pa} \in F_X^{-1}(\hat{\mathcal{O}}_X) \text{ with } (\vec{o}_{pa})_i = x_i\}$ and $\sqcap$ denotes component-wise intersection.
\end{definition}

\begin{definition}[descendant constraints]\label{def:desc-constraints}
    For SCM $\model$, target variable $Y \in \mathbf{V}$, target $\hat{\mathcal{Y}} \subseteq \mathcal{Y}$, variable $X \in \mathbf{V}$, and intervention set $\vec{S} \subseteq \mathbf{V}$, the \textbf{descendant constraints} $\hat{\mathcal{O}}^{\downarrow}_{\model,do(\vec{S})}(X,\hat{\mathcal{Y}})$ are:
    \begin{align}
        \hat{\mathcal{O}}^{\downarrow}_{\model,do(\vec{S})}(X,\hat{\mathcal{Y}}) :=
        \begin{cases}
            \{\mathcal{X}_i \mid X_i \in (\mathbf{U} \cup \mathbf{V}) \setminus Y\} \cup \{\hat{\mathcal{Y}}\} & \text{if } X = Y, \\
            \{\mathcal{X}_i \mid X_i \in \mathbf{U} \cup \mathbf{V}\} & \text{if } ch(X) = \emptyset, \\
            \displaystyle\bigsqcap_{Z \in ch(X) \setminus \vec{S}} \hat{\mathcal{O}}^{pa}_\model(Z, \hat{\mathcal{O}}^{\downarrow}_{\model,do(\vec{S})}(Z,\hat{\mathcal{Y}})) & \text{otherwise.}
        \end{cases}
    \end{align}
\end{definition}

\begin{definition}[global constraints]\label{def:global-constraints}
    The \textbf{global constraints} $\hat{\mathcal{O}}^{\mathrm{con}}_{\model,do(\vec{S})}(\hat{\mathcal{Y}}) := \bigsqcap_{R \in \mathbf{U} \cup \vec{S}} \hat{\mathcal{O}}^{\downarrow}_{\model,do(\vec{S})}(R, \hat{\mathcal{Y}})$ are the component-wise intersection of descendant constraints across all roots of $\model_{do(\vec{S})}$ (Definition~\ref{def:global-targets}).
\end{definition}

\begin{remark}[outer approximation]\label{rem:outer-approx}
    The projection $(F_X^{-1}(\hat{\mathcal{O}}_X))_i$ preserves \emph{marginal} feasibility but not \emph{joint} feasibility: every $x_i \in (F_X^{-1}(\hat{\mathcal{O}}_X))_i$ admits some complete parent configuration $\vec{o}_{pa}$ with $(\vec{o}_{pa})_i = x_i$, but the valid configurations for different parent components may be incompatible. As a consequence, the global constraints satisfy $\hat{\mathcal{O}}^{\mathrm{con}}_{\model,do(\vec{S})}(\hat{\mathcal{Y}}) \supseteq \hat{\mathcal{O}}_{\model,do(\vec{S})}(\hat{\mathcal{Y}})$: they are an outer approximation of the exact global targets. False positives arise precisely when some mechanism on the path from $X$ to $Y$ has a non-product-structured preimage (e.g., XOR: $F_C(A,B) = A \oplus B$ with $\hat{\mathcal{C}} = \{1\}$ gives projections $\hat{\mathcal{A}} = \hat{\mathcal{B}} = \{0,1\}$, yet $(0,0)$ passes both checks but violates the target). For mechanisms with product-structured preimages (e.g., AND, OR, or any parent-separable function), the approximation is exact.
\end{remark}

The global constraints can replace the global targets in Algorithm~\ref{alg:find-causes} as an efficient \emph{screening} step for a partition $(\vec{X}, \vec{Z})$ of a node $\vec{S}$: if $\vec{u}(U) \notin (\hat{\mathcal{O}}^{\mathrm{con}}_{\model,do(\vec{S})}(\hat{\mathcal{Y}}))_U$ for some $U \in \mathbf{U}$, or $\world_{\vec{u}}(Z_j) \notin (\hat{\mathcal{O}}^{\mathrm{con}}_{\model,do(\vec{S})}(\hat{\mathcal{Y}}))_{Z_j}$ for some $Z_j \in \vec{Z}$, then no witness exists (sound exclusion). If all component checks pass, a forward-simulation verification $\model_{do(\vec{X}=\vec{x}', \vec{Z}=\world_{\vec{u}}(\vec{Z}))}, \vec{u} \models (Y \in \hat{\mathcal{Y}})$ confirms or refutes the candidate; this step needs a candidate value $\vec{x}'$ and therefore, in general, reintroduces a search over alternative values. The screen thus trades exactness for space: with $D := \max_i |\mathcal{X}_i|$, the constraint collection needs $O((|\mathbf{U}| + |\mathbf{V}|) \cdot D)$ space, and a backward pass over the mechanism tables takes $O(|\mathbf{V}| \cdot D^{\Delta+1})$ time, with $\Delta$ the largest number of endogenous parents (Appendix~\ref{app:complexity-setup}), avoiding the potentially exponential joint target representation.

\section{Full Complexity Analysis}\label{app:complexity}

We give the complete cost analysis of Algorithm~\ref{alg:find-causes} across
three parameterisations of the search space.
Throughout, write $a := |anc^\mathbf{V}_\model(Y)|$,
$D := \max_i |\mathcal{X}_i|$,
$\Delta := \max_X |pa^\mathbf{V}_\model(X)|$, $p := |\mathbf{V}|$, and $o := |\mathbf{U}|$.

\subsection{Setup: Base Costs}\label{app:complexity-setup}

Two components determine the per-node cost in Phase~2 and are
shared by all three parameterisations below.

\paragraph{Target representation.}
The global targets $\hat{\mathcal{O}}_{\model,do(\vec{S})}(\hat{\mathcal{Y}})$
are subsets of the joint domain
$\vec{\mathcal{O}} = \prod_{i} \mathcal{X}_i$,
which has cardinality $|\vec{\mathcal{O}}| \leq D^{o+p}$.
With an explicit (enumerated) set representation, storing a single global target
requires $O(D^{o+p})$ space; this is the fundamental bottleneck of exact computation
and motivates the factored outer approximation of
Appendix~\ref{app:constraints}.
The required preimage projections $(F_{X_i}^{-1}(x_i))_j$
can be precomputed
once in $O(p \cdot D^{\Delta+1})$ time and stored as lookup tables of total size
$O(p \cdot D^{\Delta+1})$.
Phase~1 performs one bottom-up pass over the $p$ endogenous variables at
worst-case cost $O(p \cdot D^{o+p} \cdot \Delta)$.

\paragraph{Per-node target recomputation.}
Per-node recomputation in Phase~2 (Remark~\ref{rem:reuse}) costs
\begin{equation*}
    T_{\mathrm{target}}
    = O\!\left(|anc_\model(\vec{S})| \cdot |\vec{\mathcal{O}}| \cdot \Delta\right)
    \leq O\!\left(a \cdot D^{o+p} \cdot \Delta\right).
\end{equation*}
The mechanism class modulates $T_{\mathrm{target}}$ substantially:
for mechanisms with product-structured preimages (AND, OR, linear, monotone),
the global targets factorise and $T_{\mathrm{target}} = O(a \cdot D^{\Delta})$;
for continuous variables $\mathcal{X}_i \subseteq \mathbb{R}$,
targets are represented implicitly (polytopes, intervals, semialgebraic sets)
at cost $T_{\mathrm{target}} = O(a \cdot \rho)$,
where $\rho$ is the cost of a
single preimage computation:
$\rho = O(p)$ for linear mechanisms (preimage is a hyperplane),
$\rho = O(1)$ for monotone mechanisms (marginal interval arithmetic, exact via
Appendix~\ref{app:constraints}), and
$\rho = O(\log(1/\varepsilon))$ for general smooth mechanisms (bisection).
In all cases the logical structure of Algorithm~\ref{alg:find-causes} and the
guarantees of Theorem~\ref{thm:sound-complete} are unaffected; only the
membership subroutine requires adaptation to the mechanism class.

\subsection{Enumeration Cost under Three Parameterisations}\label{app:complexity-cases}

The three parameterisations restrict the search space of the ancestral map
in qualitatively different ways; the per-node cost $T_{\mathrm{target}}$
from Section~\ref{app:complexity-setup} applies uniformly in each case.

\paragraph{(i) Unrestricted.}
The ancestral map has at most $2^a$ nodes.
At a node $\vec{S}$ of size $s$, the algorithm tests $2^s - 1$ subsets with exactly
one membership query each (Corollary~\ref{cor:witness-existence}).
The total number of membership queries is bounded by
\[
    \sum_{s=1}^{a} \binom{a}{s} \cdot (2^s - 1)
    <
    \sum_{s=0}^{a} \binom{a}{s} \cdot 2^s
    =
    3^a.
\]
The total enumeration cost is therefore exponential in $a$ in the worst case.

\paragraph{(ii) Bounded depth $d$.}
Restricting traversal to nodes reachable within $d$ derivation steps from
$\{Y\}$, the analysis of case~(i) applies with $a$ replaced by
$a_d \leq 2\Delta^d$ (the number of ancestors within $d$ hops for bounded
in-degree $\Delta$), yielding a membership-query bound of $3^{2\Delta^d}$.
At depth $d = 1$ the search is restricted to direct parents of $Y$ and
performs at most $3^\Delta$ membership queries.

\paragraph{(iii) Bounded intervention size $|\vec{S}| \leq s$.}
Restricting to nodes of size at most $s$ yields at most
$\sum_{k=1}^{s}\binom{a}{k} = O(a^s)$ nodes,
with at most $2^s - 1$ membership queries per node.
For product-structured mechanisms the total cost is
\begin{equation*}
    O\!\left(a^s \cdot 2^s \cdot p \cdot D^{\Delta}\right),
\end{equation*}
which is polynomial in $p$, $a$, and $D$ for fixed $s$ and $\Delta$.
For continuous or general mechanism classes, substituting
$T_{\mathrm{target}} = O(a \cdot \rho)$ gives the unified bound
\begin{equation*}
    O\!\left(a^s \cdot 2^s \cdot (p \cdot \rho + T_{\mathrm{mem}})\right),
\end{equation*}
where $T_{\mathrm{mem}}$ is the cost of a single membership query.
Phase~3 verification requires checking at most $2^s$ strict subsets per
candidate.

\paragraph{Expected-case behaviour.}
The worst-case bound $|\mathcal{A}_\model(Y)| \leq 2^a$ is loose in practice:
ancestor subsets violating the path-disjointness condition of
Proposition~\ref{prop:ancestor-reachability} are typically not reached, and
those that are tend to fail the node-level pre-check
(Appendix~\ref{app:pruning}(ii)) in overdetermined regimes.
Empirically (Appendix~\ref{app:experiments}, Figure~\ref{fig:app-pruning}),
the combined pruning of Remark~\ref{rem:pruning}(ii) and (iii) eliminates
on average over $97\%$ of witness-map nodes for $n \geq 10$ at densities
$d \geq 0.5$ and $73$--$94\%$ at $d = 0.25$.
Under bounded in-degree $\Delta$ and bounded tree-width $\tau$ of
$\mathcal{G}_\model$, paths from ancestors to $Y$ are more constrained, so
fewer subsets satisfy the path-disjointness condition; while this does not
change the $O(a^s)$ asymptotic for case~(iii), the implicit constant
decreases with $\tau$.
A formal expected-case analysis under random-DAG models is beyond the
present scope.

\section{Phase~2 Pruning Strategies}\label{app:pruning}

The three strategies are stated in Remark~\ref{rem:pruning}; we give the
formal justifications here, including how each strategy behaves under
continuous variable domains.

\paragraph{(i) Size-first traversal.}
Because the ancestral map $\mathcal{A}_\model(Y)$ is a finite DAG and actual
causes form an antichain under $\subsetneq$ (AC3), processing nodes in
non-decreasing order of $|\vec{S}|$ means that whenever a cause $\vec{X}' \in \mathcal{C}$
is confirmed, every node $\vec{S}'$ with $\vec{X}' \subsetneq \vec{S}'$ has not yet been processed,
so strategy~(iii) achieves its maximal pruning effect.
This strategy operates purely on the lattice structure of the ancestral map and
carries over unchanged to continuous variable domains.
\emph{Cost saving.} Combined with strategy~(iii), size-first traversal makes
Phase~3 redundant: the antichain invariant of $\mathcal{C}$ is maintained
\emph{during} Phase~2, eliminating the $O(|\mathcal{C}|^2 \cdot s_{\max})$
post-filter pairwise antichain comparisons (where $s_{\max} = \max_{\vec{X} \in \mathcal{C}} |\vec{X}|$)
at zero per-query overhead.

\paragraph{(ii) Node-level pre-check.}
For any non-empty partition $\vec{X} \cup \vec{Z} = \vec{S}$, the existence of $\vec{o} \in \hat{\mathcal{O}}_{\model,do(\vec{S})}(\hat{\mathcal{Y}})$ with $\vec{o}_\mathbf{U} = \vec{u}$ and $\vec{o}_{\vec{Z}} = \world_{\vec{u}}(\vec{Z})$ implies in particular the existence of $\vec{o}' \in \hat{\mathcal{O}}_{\model,do(\vec{S})}(\hat{\mathcal{Y}})$ with $\vec{o}'_\mathbf{U} = \vec{u}$ (taking $\vec{o}' := \vec{o}$).
Contrapositively, failure of the latter weaker condition is a \emph{sufficient} reason to skip the entire node, since no partition $(\vec{X}, \vec{Z})$ of $\vec{S}$ can pass the membership test.
In redundantly overdetermined settings (e.g.\ $F_Y = A \vee C$ with independent
$A = C = 1$), this eliminates every root-adjacent node whose intervention set
is a strict subset of $pa^\mathbf{V}_\model(Y)$, while
$pa^\mathbf{V}_\model(Y)$ itself typically passes the check.
\emph{Cost saving.} Each failed node $\vec{S}$ skips all $2^{|\vec{S}|}-1$ partition
queries at the cost of one membership query, a net saving of $2^{|\vec{S}|}-2$
queries. Across the witness map, the Phase~2 query count drops from
$\sum_{\vec{S}} (2^{|\vec{S}|}-1)$ to
$\mathcal{N} + \sum_{\vec{S} \text{ passes}} (2^{|\vec{S}|}-1)$,
where $\mathcal{N}$ is the number of pre-check failures.
The savings dominate in overdetermined regimes (e.g.\ Forest-fire Case~I in
Appendix~\ref{app:example-forest-fire}, where two of three non-root nodes
fail the pre-check).

The pre-check remains formally valid in continuous domains; only its
\emph{implementation} changes, since deciding the existence of $\vec{o} \in \hat{\mathcal{O}}_{\model,do(\vec{S})}(\hat{\mathcal{Y}})$ with $\vec{o}_\mathbf{U} = \vec{u}$
becomes a real-valued feasibility problem rather than a hash-table lookup.
The constraint-based outer approximation of
Appendix~\ref{app:constraints}
provides an efficient and often exact screening step in all three
mechanism classes:
for \emph{monotone mechanisms}, marginal preimage projections are closed intervals
and the check reduces to component-wise interval comparisons, which are exact
(Remark~\ref{rem:outer-approx});
for \emph{linear-Gaussian mechanisms}, the check reduces to a single LP
feasibility query;
for \emph{general smooth mechanisms}, bisection or gradient-based root-finding
locates a witness $\vec{x}'$ or certifies non-existence.

\paragraph{(iii) Superset pruning.}
If $\vec{X}' \subsetneq \vec{X}$ and $\vec{X}' \in \mathcal{C}$, then by (AC3) $\vec{X}$ cannot be
minimal and would be discarded in Phase~3 regardless.
Maintaining $\mathcal{C}$ as an antichain under $\subsetneq$ throughout Phase~2
therefore makes Phase~3 redundant when combined with strategy~(i):
\[
    \vec{X} \in
    \bigl\{\emptyset \neq \vec{X} \subseteq \vec{S}
    \bigm|
    \nexists\, \vec{X}' \in \mathcal{C} \colon \vec{X}' \subsetneq \vec{X}\bigr\}.
\]
Like strategy~(i), superset pruning operates purely on the antichain structure
of $\mathcal{C}$ and carries over unchanged to continuous variable domains.
\emph{Cost saving.} For each confirmed cause $\vec{X}' \in \mathcal{C}$ of size $k$
and each later-processed node $\vec{S} \supseteq \vec{X}'$, the
$2^{|\vec{S}|-k}-1$ proper supersets of $\vec{X}'$ within $\vec{S}$
(every set extending $\vec{X}'$ by a non-empty subset of $\vec{S} \setminus \vec{X}'$)
are skipped before the membership query.
The savings compound multiplicatively in regimes with many small confirmed
causes.

\section{Extended Experimental Results}\label{app:experiments}

\subsection{Setup}

All experiments use randomly generated Boolean SCMs.
The causal DAG is sampled from an acyclic Erd\H{o}s-R\'enyi model: each
ordered pair of nodes is connected by a directed edge independently with
probability $d$.
Every structural equation $F_X$ is a random AND/OR gate applied to the parents
of $X$, with each parent input independently negated with probability $0.5$.
We vary the number of endogenous nodes $n \in \{3, 4, \ldots, 15\}$ and edge
density $d \in \{0.25, 0.50, 0.75\}$, drawing $10$ independent seeds per
$(n,d)$ configuration ($390$ instances in total).
The target variable $Y$ is chosen as the deepest node in the graph; the
environment $\vec{u}$ is drawn uniformly at random, and target event
$\mathcal{Y}^* = \{1\}$ is used throughout.
TrAC is run with all three pruning strategies of Remark~\ref{rem:pruning} implemented
(size-first traversal, node-level pre-check, superset pruning).
Individual runs are sequentially benchmarked on a single CPU core of an AMD Ryzen Threadripper 1900X 8-Core Processor with 32GB RAM.

\textbf{Baselines.}
\citet{Reyd2025} identify
a single cause via forward simulation with \emph{beam search} and
\emph{ISI} by sampling alternative variable values.
This method supports non-Boolean and black-box SCMs but provides no soundness or
completeness guarantee.
\citet{Ibrahim2020} encode the HP
conditions as propositional formulae and apply three solving strategies: Brute
Force (BF), SAT, and MaxSAT. These methods are restricted to Boolean SCMs.
ILP/ILP-WHY strategies are excluded as they require a Gurobi licence.
Both baselines target a single (minimal) cause and are not designed for full
enumeration; the runtime comparison is therefore indicative rather than strictly
analogous to TrAC.

\subsection{Results}

\begin{figure}[h]
    \centering
    \includegraphics[width=0.98\linewidth]{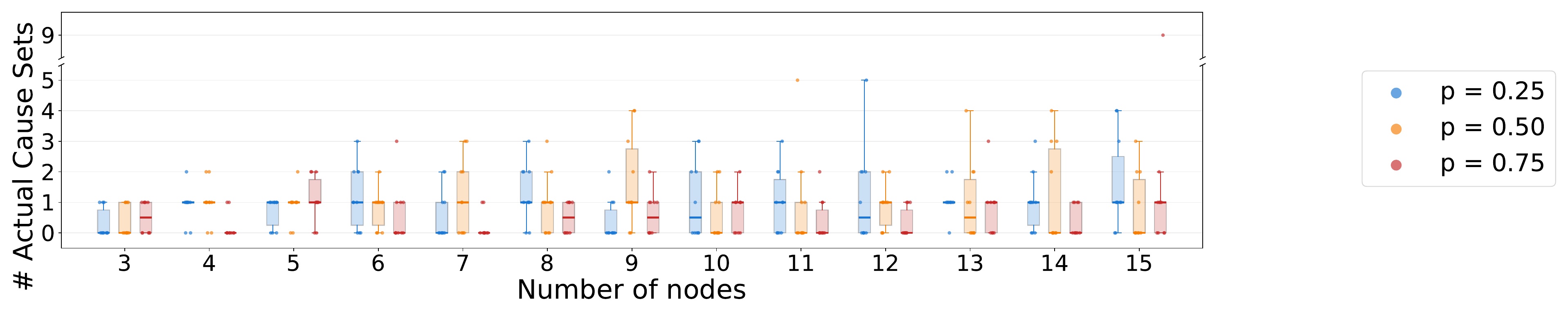}
    \caption{\textbf{Number of actual cause sets found by TrAC.}
    Box-and-whisker across 10 seeds (median line, $[25,75]$-IQR box, whiskers, outlier markers).
    The number of causes grows moderately with graph size and edge density.}
    \label{fig:app-n-causes}
\end{figure}

\textbf{Cause counts.}
Figure~\ref{fig:app-n-causes} shows the number of actual cause sets found by
TrAC per instance.
The number of causes grows moderately with both $n$ and $d$.
Even at moderate cause counts, a principled enumeration algorithm is
indispensable for downstream aggregation tasks such as responsibility and blame,
as the combinatorial structure of the witness map, not the number of causes
alone, drives the difficulty of complete enumeration.

\begin{figure}[h]
    \centering
    \includegraphics[width=0.9\linewidth]{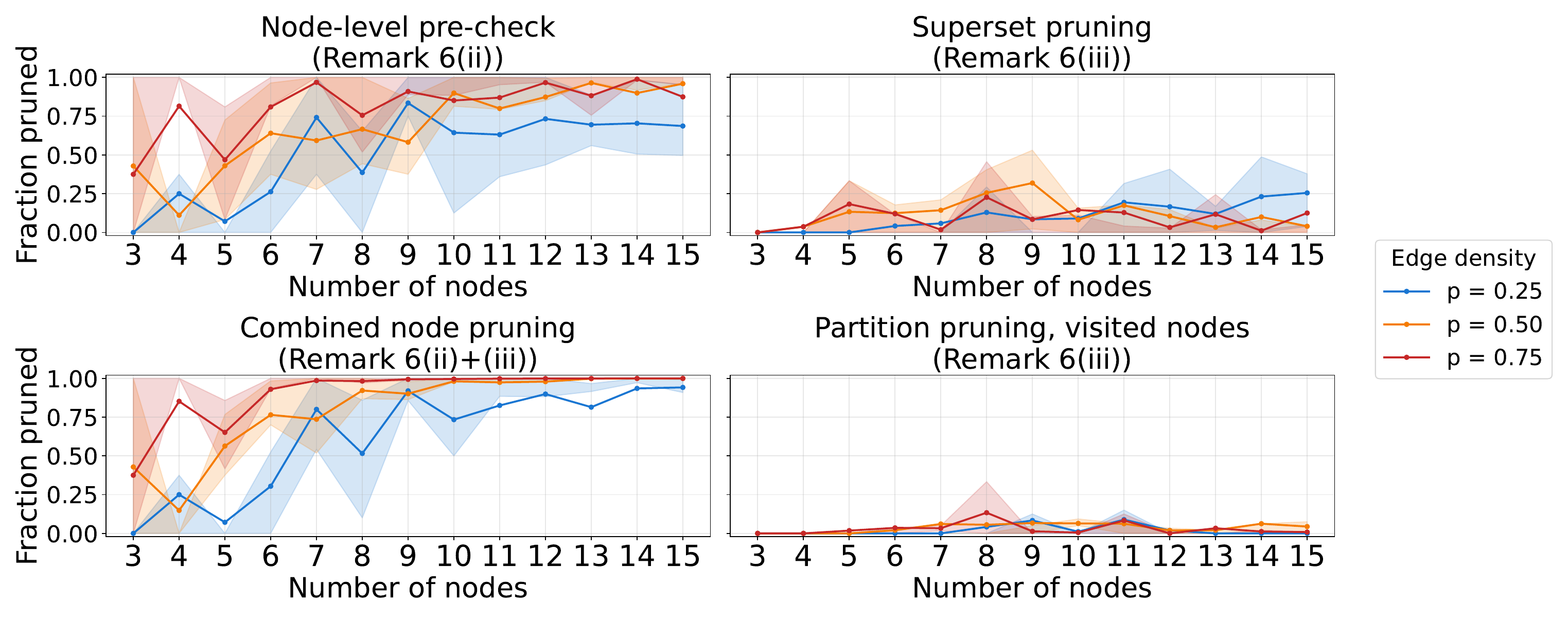}
    \caption{\textbf{Pruning rates.}
    Fraction of witness-map nodes eliminated by each strategy (mean $\pm$
    $[25,75]$-IQR), broken down by node count and density.
    The node-level pre-check dominates; combined with superset pruning, on average over $97\%$
    of nodes are eliminated for $n \geq 10$ at $d \geq 0.5$, and $73$--$94\%$ at $d = 0.25$.}
    \label{fig:app-pruning}
\end{figure}

\textbf{Pruning effectiveness.}
Figure~\ref{fig:app-pruning} reports the fraction of work eliminated by each
pruning strategy of Remark~\ref{rem:pruning}.
The node-level pre-check (strategy~(ii)) is by far the most effective: at
$n \geq 10$ it prunes more than $96\%$ of all witness-map nodes before any
partition enumeration, reaching $99.98\%$ at $n{=}15$, $d{=}0.75$.
Superset pruning (strategy~(iii)) provides additional savings at the partition
level, while size-first traversal (strategy~(i)) amplifies both by ensuring
confirmed causes are discovered as early as possible.
Together, these three strategies reduce the practical cost of Phase~2 by several
orders of magnitude relative to exhaustive traversal.

\begin{figure}[h]
    \centering
    \includegraphics[width=0.98\linewidth]{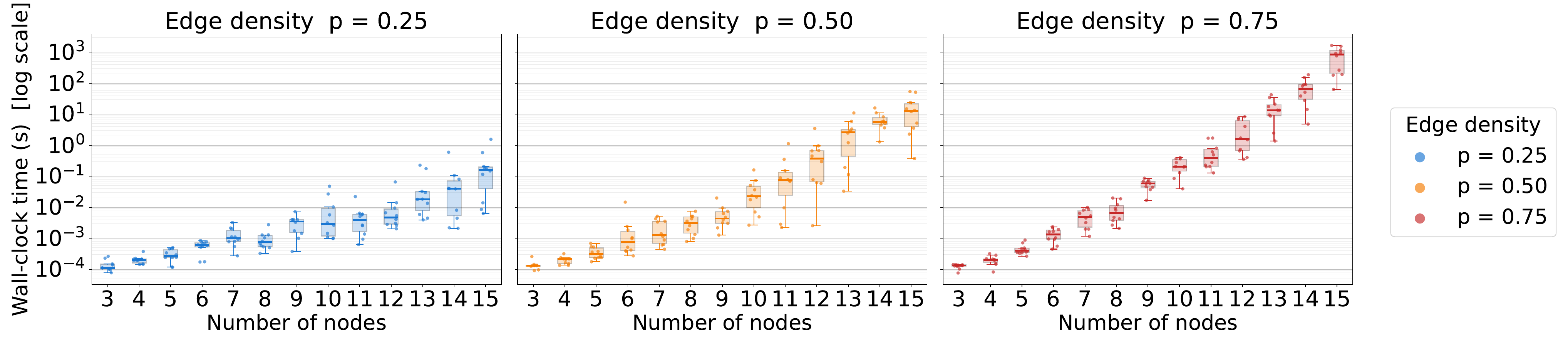}
    \caption{\textbf{TrAC wall-clock runtime on random Boolean SCMs.}
    Box-and-whisker across 10 seeds per $(n,d)$ cell (median line, $[25,75]$-IQR box, whiskers, outlier markers).
    Runtime scales steeply at high density ($d{=}0.75$) due to the larger witness
    maps induced by denser graphs.}
    \label{fig:app-trac}
\end{figure}

\textbf{TrAC standalone runtime.}
Figure~\ref{fig:app-trac} presents TrAC's wall-clock runtime for each $(n, d)$
configuration.
Runtime grows with both $n$ and $d$: at $n{=}15$ the mean ranges from
$0.30\,\mathrm{s}$ ($d{=}0.25$) to $18\,\mathrm{s}$ ($d{=}0.50$) and
$773\,\mathrm{s}$ ($d{=}0.75$), with substantial inter-seed variance at high
density visible in the widening IQR bands.

\begin{figure}[h]
    \centering
    \includegraphics[width=0.98\linewidth]{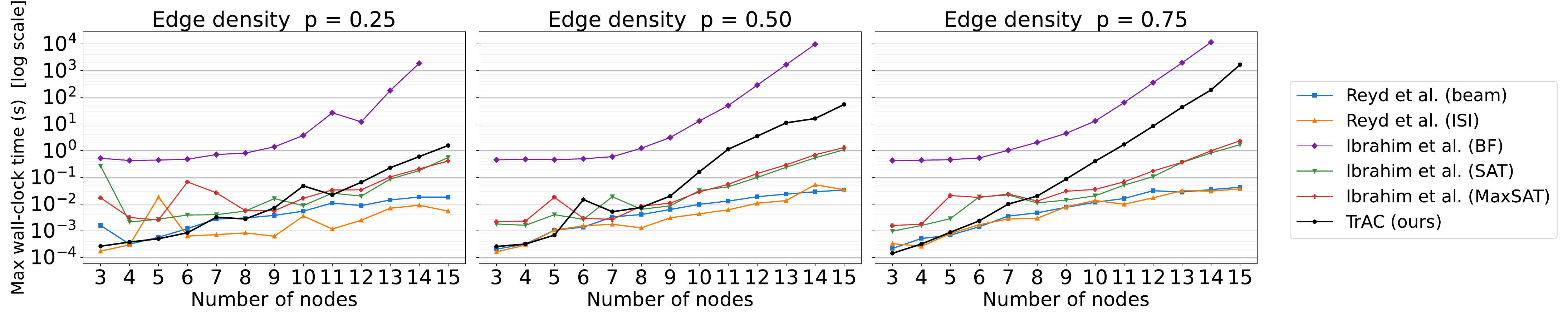}
    \caption{\textbf{Worst-case runtime comparison.}
    Maximum observed runtime per $(n,d)$ cell.
    HP2SAT BF was excluded at $n{=}15$ due to exploding runtime and memory
    requirements; TrAC completes every instance, with a worst-case runtime of
    approximately $1{,}600\,\mathrm{s}$ at $n{=}15$, $d{=}0.75$.}
    \label{fig:app-compare-max}
\end{figure}

\textbf{Worst-case runtime.}
Figure~\ref{fig:app-compare-max} shows the maximum runtime per $(n, d)$ cell,
highlighting the tail behaviour of each method.
HP2SAT BF exhibits the most extreme worst cases and was excluded at $n{=}15$ due
to exploding runtime and memory requirements.
TrAC's worst-case runtime grows with $n$ and $d$ but remains finite across all
tested configurations.

\section{Roadmap and Open Problems}\label{app:roadmap}

\textbf{Context-specific independence and pruning.}
The node-level pre-check (Remark~\ref{rem:pruning}(ii)) has a precise algebraic
interpretation that connects to context-specific independence (CSI)
\citep{Boutilier1996,Zhang1999}.
Specifically, the parent target operator (Definition~\ref{def:parent-targets})
filters joint configurations by checking, for each child $X_i$ along the
backward path, whether the parent tuple of a configuration lies in the joint
preimage $F_{X_i}^{-1}(\vec{o}_i)$ at that configuration's child value
$\vec{o}_i$.
The \emph{outer approximation} of Appendix~\ref{app:constraints} relaxes this
joint check to a set-level operation that tracks per-variable marginal subsets
via $(F_X^{-1}(\hat{\mathcal{O}}_X))_i$, and thereby already realises the
marginal (component-wise) version of CSI-based pruning \citep{Pensar2015},
making the precision gap between the exact and approximate algorithms
transparent (see Remark~\ref{rem:outer-approx}).
A formal characterisation of the conditions under which pre-check failure
coincides with graph-theoretic CSI criteria remains an open problem.

\textbf{Compact representations of global targets.}
Global targets are subsets of the joint domain $\vec{\mathcal{O}}$, requiring
$O(D^{o+p})$ space in the worst case.
For product-structured mechanisms (AND, OR, linear, monotone), targets factorise
and storage reduces to $O(p \cdot D^\Delta)$; for Boolean mechanisms, binary
decision diagrams (BDDs) support polynomial-time set operations (intersection, union, complement) on the target sets, though BDD size can itself be exponential in the number of variables in the worst case; for linear mechanisms, polytopes in half-space representation reduce
membership to LP feasibility.
A systematic characterisation of the broadest mechanism classes admitting compact representations would eliminate the exponential storage bottleneck and
enable the algorithm to scale to large acyclic models.

\textbf{Responsibility, blame, harm, and risk.}
Algorithm~\ref{alg:find-causes} returns all actual causes for a fixed environment
$\vec{u}$; responsibility and blame scores \citep{Chockler2004} and causal harm
assessments \citep{BeckersChocklerHalpern2022} are defined as weighted aggregates
over actual causes and contingency sets, so the witness map provides precisely the
missing computational substrate for these downstream quantities.
Extending the algorithm to a full distribution $P(\mathbf{U})$ amounts to
marginalising over environments: for a fixed variable set $\vec{X}$ and value $\vec{x}$,
the \emph{probability that $(\vec{X}{=}\vec{x})$ is an actual cause of $(Y \in \mathcal{Y}^*)$}
equals the $P(\mathbf{U})$-measure of environments $\vec{u}$ satisfying
$\world_{\vec{u}}(\vec{X}) = \vec{x}$, $\model, \vec{u} \models (Y \in \mathcal{Y}^*)$, and
passing the membership test for $\vec{X}$ (with some contingency $\vec{Z}$) in Phase~2 with
$\vec{X}$ retained after the minimality filter of Phase~3.
This probabilistic lifting connects directly to the framework of probabilistic
actual causation \citep{FentonGlynn2017} and to nondeterministic extensions
\citep{Beckers2025}: in both cases the witness map structure is preserved,
with the binary membership predicate replaced by a measure over the set of
witnessing environments.

\clearpage
\section*{NeurIPS Paper Checklist}

\begin{enumerate}

\item {\bf Claims}
    \item[] Question: Do the main claims made in the abstract and introduction accurately reflect the paper's contributions and scope?
    \item[] Answer: \answerYes{}
    \item[] Justification: The paper presents a sound and complete algorithm for tracing all actual causes. Individual contributions regarding witness targets are given in Section~\ref{sec:witness-targets}, specifically Definitions~\ref{def:parent-targets}--\ref{def:global-targets}, Theorem~\ref{thm:witness-target}, Corollary~\ref{cor:witness-existence}. Witness maps are outlined in Section~\ref{sec:witness-maps}, specifically Definitions~\ref{def:anc-grammar}-\ref{def:witness-map}, Propositions~\ref{prop:grammar-equiv}-\ref{prop:ancestor-reachability}. The Tracing Actual Causes (TrAC) algorithm is introduced in Section~\ref{sec:algorithm} and Algorithm~\ref{alg:find-causes} with soundness and completeness proof (Theorem~\ref{thm:sound-complete}) and complexity analysis (Sec.~\ref{app:complexity}). Finally, experiments and baseline comparisons are presented in Sec.~\ref{sec:experiments} and Appendix~\ref{app:experiments}.
    \item[] Guidelines:
    \begin{itemize}
        \item The answer \answerNA{} means that the abstract and introduction do not include the claims made in the paper.
        \item The abstract and/or introduction should clearly state the claims made, including the contributions made in the paper and important assumptions and limitations. A \answerNo{} or \answerNA{} answer to this question will not be perceived well by the reviewers.
        \item The claims made should match theoretical and experimental results, and reflect how much the results can be expected to generalize to other settings.
        \item It is fine to include aspirational goals as motivation as long as it is clear that these goals are not attained by the paper.
    \end{itemize}

\item {\bf Limitations}
    \item[] Question: Does the paper discuss the limitations of the work performed by the authors?
    \item[] Answer: \answerYes{}
    \item[] Justification: Verifying a single candidate cause under the modified Halpern--Pearl definition is $\mathsf{D}_1^P$-complete~\citep{Aleksandrowicz2014,Halpern2015}, so exponential worst-case cost is inherent to exact enumeration. Runtime and storage limitations of TrAC are analysed in Sec.~\ref{sec:algorithm}, Appendix~\ref{app:complexity} and the limitations paragraph of Sec.~\ref{sec:conclusion}.
    \item[] Guidelines:
    \begin{itemize}
        \item The answer \answerNA{} means that the paper has no limitation while the answer \answerNo{} means that the paper has limitations, but those are not discussed in the paper.
        \item The authors are encouraged to create a separate ``Limitations'' section in their paper.
        \item The paper should point out any strong assumptions and how robust the results are to violations of these assumptions (e.g., independence assumptions, noiseless settings, model well-specification, asymptotic approximations only holding locally). The authors should reflect on how these assumptions might be violated in practice and what the implications would be.
        \item The authors should reflect on the scope of the claims made, e.g., if the approach was only tested on a few datasets or with a few runs. In general, empirical results often depend on implicit assumptions, which should be articulated.
        \item The authors should reflect on the factors that influence the performance of the approach. For example, a facial recognition algorithm may perform poorly when image resolution is low or images are taken in low lighting. Or a speech-to-text system might not be used reliably to provide closed captions for online lectures because it fails to handle technical jargon.
        \item The authors should discuss the computational efficiency of the proposed algorithms and how they scale with dataset size.
        \item If applicable, the authors should discuss possible limitations of their approach to address problems of privacy and fairness.
        \item While the authors might fear that complete honesty about limitations might be used by reviewers as grounds for rejection, a worse outcome might be that reviewers discover limitations that aren't acknowledged in the paper. The authors should use their best judgment and recognize that individual actions in favor of transparency play an important role in developing norms that preserve the integrity of the community. Reviewers will be specifically instructed to not penalize honesty concerning limitations.
    \end{itemize}

\item {\bf Theory assumptions and proofs}
    \item[] Question: For each theoretical result, does the paper provide the full set of assumptions and a complete (and correct) proof?
    \item[] Answer: \answerYes{}
    \item[] Justification: The paper presents a sound and complete algorithm for tracing all actual causes under the modified Halpern-Pearl definition of actual causality on acyclic SCM. (Stated in the abstract, Sec.~\ref{sec:intro} and Sec.~\ref{sec:finding-causes}). Individual definitions regarding witness targets are given in Section~\ref{sec:witness-targets}, Definitions~\ref{def:parent-targets}--\ref{def:global-targets}, Theorem~\ref{thm:witness-target}, Corollary~\ref{cor:witness-existence}. Witness maps are outlined in Section~\ref{sec:witness-maps}, specifically Definitions~\ref{def:anc-grammar}-\ref{def:witness-map}, Propositions~\ref{prop:grammar-equiv}-\ref{prop:ancestor-reachability}. The complete Tracing Actual Causes (TrAC) algorithm is introduced in Section~\ref{sec:algorithm} and Algorithm~\ref{alg:find-causes} with soundness and completeness proof (Theorem~\ref{thm:sound-complete}) and complexity analysis (Sec.~\ref{app:complexity}).
    \item[] Guidelines:
    \begin{itemize}
        \item The answer \answerNA{} means that the paper does not include theoretical results.
        \item All the theorems, formulas, and proofs in the paper should be numbered and cross-referenced.
        \item All assumptions should be clearly stated or referenced in the statement of any theorems.
        \item The proofs can either appear in the main paper or the supplemental material, but if they appear in the supplemental material, the authors are encouraged to provide a short proof sketch to provide intuition.
        \item Inversely, any informal proof provided in the core of the paper should be complemented by formal proofs provided in appendix or supplemental material.
        \item Theorems and Lemmas that the proof relies upon should be properly referenced.
    \end{itemize}

    \item {\bf Experimental result reproducibility}
    \item[] Question: Does the paper fully disclose all the information needed to reproduce the main experimental results of the paper to the extent that it affects the main claims and/or conclusions of the paper (regardless of whether the code and data are provided or not)?
    \item[] Answer: \answerYes{}
    \item[] Justification: Experimental details on tested parameter configurations, graph and data sampling, as well as repetitions are described in Sec.~\ref{sec:experiments} and Appendix~\ref{app:experiments}. All code is available at \url{https://github.com/MoritzWillig/tracingActualCauses}.
    \item[] Guidelines:
    \begin{itemize}
        \item The answer \answerNA{} means that the paper does not include experiments.
        \item If the paper includes experiments, a \answerNo{} answer to this question will not be perceived well by the reviewers: Making the paper reproducible is important, regardless of whether the code and data are provided or not.
        \item If the contribution is a dataset and\slash or model, the authors should describe the steps taken to make their results reproducible or verifiable.
        \item Depending on the contribution, reproducibility can be accomplished in various ways. For example, if the contribution is a novel architecture, describing the architecture fully might suffice, or if the contribution is a specific model and empirical evaluation, it may be necessary to either make it possible for others to replicate the model with the same dataset, or provide access to the model. In general. releasing code and data is often one good way to accomplish this, but reproducibility can also be provided via detailed instructions for how to replicate the results, access to a hosted model (e.g., in the case of a large language model), releasing of a model checkpoint, or other means that are appropriate to the research performed.
        \item While NeurIPS does not require releasing code, the conference does require all submissions to provide some reasonable avenue for reproducibility, which may depend on the nature of the contribution. For example
        \begin{enumerate}
            \item If the contribution is primarily a new algorithm, the paper should make it clear how to reproduce that algorithm.
            \item If the contribution is primarily a new model architecture, the paper should describe the architecture clearly and fully.
            \item If the contribution is a new model (e.g., a large language model), then there should either be a way to access this model for reproducing the results or a way to reproduce the model (e.g., with an open-source dataset or instructions for how to construct the dataset).
            \item We recognize that reproducibility may be tricky in some cases, in which case authors are welcome to describe the particular way they provide for reproducibility. In the case of closed-source models, it may be that access to the model is limited in some way (e.g., to registered users), but it should be possible for other researchers to have some path to reproducing or verifying the results.
        \end{enumerate}
    \end{itemize}

\item {\bf Open access to data and code}
    \item[] Question: Does the paper provide open access to the data and code, with sufficient instructions to faithfully reproduce the main experimental results, as described in supplemental material?
    \item[] Answer: \answerYes{}
    \item[] Justification: All code and recorded runtime results are available at \url{https://github.com/MoritzWillig/tracingActualCauses}. The included README contains all instructions to run the benchmark and plot the figures.
    \item[] Guidelines:
    \begin{itemize}
        \item The answer \answerNA{} means that paper does not include experiments requiring code.
        \item Please see the NeurIPS code and data submission guidelines (\url{https://neurips.cc/public/guides/CodeSubmissionPolicy}) for more details.
        \item While we encourage the release of code and data, we understand that this might not be possible, so \answerNo{} is an acceptable answer. Papers cannot be rejected simply for not including code, unless this is central to the contribution (e.g., for a new open-source benchmark).
        \item The instructions should contain the exact command and environment needed to run to reproduce the results. See the NeurIPS code and data submission guidelines (\url{https://neurips.cc/public/guides/CodeSubmissionPolicy}) for more details.
        \item The authors should provide instructions on data access and preparation, including how to access the raw data, preprocessed data, intermediate data, and generated data, etc.
        \item The authors should provide scripts to reproduce all experimental results for the new proposed method and baselines. If only a subset of experiments are reproducible, they should state which ones are omitted from the script and why.
        \item At submission time, to preserve anonymity, the authors should release anonymized versions (if applicable).
        \item Providing as much information as possible in supplemental material (appended to the paper) is recommended, but including URLs to data and code is permitted.
    \end{itemize}

\item {\bf Experimental setting/details}
    \item[] Question: Does the paper specify all the training and test details (e.g., data splits, hyperparameters, how they were chosen, type of optimizer) necessary to understand the results?
    \item[] Answer: \answerYes{}
    \item[] Justification: Experimental details on tested parameter configurations, graph and data sampling, as well as repetitions are described in Sec.~\ref{sec:experiments} and Appendix~\ref{app:experiments}. Baselines algorithms are tested as provided by their respective code repositories.
    \item[] Guidelines:
    \begin{itemize}
        \item The answer \answerNA{} means that the paper does not include experiments.
        \item The experimental setting should be presented in the core of the paper to a level of detail that is necessary to appreciate the results and make sense of them.
        \item The full details can be provided either with the code, in appendix, or as supplemental material.
    \end{itemize}

\item {\bf Experiment statistical significance}
    \item[] Question: Does the paper report error bars suitably and correctly defined or other appropriate information about the statistical significance of the experiments?
    \item[] Answer: \answerYes{}
    \item[] Justification: Each $(n,d)$ configuration is evaluated on 10 independently sampled instances (random graph, structural equations and environment), so the reported variability is across instances. Figs.~\ref{fig:trac-compare} and~\ref{fig:app-pruning} show the mean with $[25,75]$-IQR bands, Figs.~\ref{fig:app-trac} and~\ref{fig:app-n-causes} box-and-whisker plots (median, IQR, whiskers, outliers), and Fig.~\ref{fig:app-compare-max} the worst case per configuration.
    \item[] Guidelines:
    \begin{itemize}
        \item The answer \answerNA{} means that the paper does not include experiments.
        \item The authors should answer \answerYes{} if the results are accompanied by error bars, confidence intervals, or statistical significance tests, at least for the experiments that support the main claims of the paper.
        \item The factors of variability that the error bars are capturing should be clearly stated (for example, train/test split, initialization, random drawing of some parameter, or overall run with given experimental conditions).
        \item The method for calculating the error bars should be explained (closed form formula, call to a library function, bootstrap, etc.)
        \item The assumptions made should be given (e.g., Normally distributed errors).
        \item It should be clear whether the error bar is the standard deviation or the standard error of the mean.
        \item It is OK to report 1-sigma error bars, but one should state it. The authors should preferably report a 2-sigma error bar than state that they have a 96\% CI, if the hypothesis of Normality of errors is not verified.
        \item For asymmetric distributions, the authors should be careful not to show in tables or figures symmetric error bars that would yield results that are out of range (e.g., negative error rates).
        \item If error bars are reported in tables or plots, the authors should explain in the text how they were calculated and reference the corresponding figures or tables in the text.
    \end{itemize}

\item {\bf Experiments compute resources}
    \item[] Question: For each experiment, does the paper provide sufficient information on the computer resources (type of compute workers, memory, time of execution) needed to reproduce the experiments?
    \item[] Answer: \answerYes{}
    \item[] Justification: Sec.~\ref{app:experiments} describes the hardware of a single AMD Ryzen Threadripper 1900X 8-Core PC with 32GB RAM (no GPU required). Runtime comparisons to baselines, individual repetition runtimes and worst-case runtimes for TrAC are reported in Figures~\ref{fig:trac-compare}, \ref{fig:app-trac} and \ref{fig:app-compare-max}.
    \item[] Guidelines:
    \begin{itemize}
        \item The answer \answerNA{} means that the paper does not include experiments.
        \item The paper should indicate the type of compute workers CPU or GPU, internal cluster, or cloud provider, including relevant memory and storage.
        \item The paper should provide the amount of compute required for each of the individual experimental runs as well as estimate the total compute.
        \item The paper should disclose whether the full research project required more compute than the experiments reported in the paper (e.g., preliminary or failed experiments that didn't make it into the paper).
    \end{itemize}

\item {\bf Code of ethics}
    \item[] Question: Does the research conducted in the paper conform, in every respect, with the NeurIPS Code of Ethics \url{https://neurips.cc/public/EthicsGuidelines}?
    \item[] Answer: \answerYes{}
    \item[] Justification: The research is purely theoretical/methodological. It involves no human subjects, no personally identifiable information or sensitive data. Experiments are conducted exclusively on synthetic data (randomly sampled causal graphs), and illustrative examples are standard, well-established toy scenarios from the causal inference literature (rock-throwing, forest fires). We anticipate no direct ethical concerns, risks of discrimination or privacy violation arising from this work.
    \item[] Guidelines:
    \begin{itemize}
        \item The answer \answerNA{} means that the authors have not reviewed the NeurIPS Code of Ethics.
        \item If the authors answer \answerNo, they should explain the special circumstances that require a deviation from the Code of Ethics.
        \item The authors should make sure to preserve anonymity (e.g., if there is a special consideration due to laws or regulations in their jurisdiction).
    \end{itemize}

\item {\bf Broader impacts}
    \item[] Question: Does the paper discuss both potential positive societal impacts and negative societal impacts of the work performed?
    \item[] Answer: \answerYes{}
    \item[] Justification: Actual causation underpins definitions of legal responsibility, moral blame, and the assessment of causal harm. Previous works in this area primarily focused on finding single smallest cause sets, blocking these tasks that either depend on finding \emph{all} actual causes in a given setting or assume a particular case was already established a-priori. By enumerating all actual causes, TrAC may thus advance reasoning in all of these areas. Conversely, automated cause enumeration could be misused to attribute blame or responsibility mechanically; its output is only as reliable as the specified causal model, and conclusions drawn from a misspecified model can be misleading. TrAC should therefore support, not replace, human judgement. Relevant discussions are included in the abstract, introduction and Appendix~\ref{app:roadmap}.
    \item[] Guidelines:
    \begin{itemize}
        \item The answer \answerNA{} means that there is no societal impact of the work performed.
        \item If the authors answer \answerNA{} or \answerNo, they should explain why their work has no societal impact or why the paper does not address societal impact.
        \item Examples of negative societal impacts include potential malicious or unintended uses (e.g., disinformation, generating fake profiles, surveillance), fairness considerations (e.g., deployment of technologies that could make decisions that unfairly impact specific groups), privacy considerations, and security considerations.
        \item The conference expects that many papers will be foundational research and not tied to particular applications, let alone deployments. However, if there is a direct path to any negative applications, the authors should point it out. For example, it is legitimate to point out that an improvement in the quality of generative models could be used to generate Deepfakes for disinformation. On the other hand, it is not needed to point out that a generic algorithm for optimizing neural networks could enable people to train models that generate Deepfakes faster.
        \item The authors should consider possible harms that could arise when the technology is being used as intended and functioning correctly, harms that could arise when the technology is being used as intended but gives incorrect results, and harms following from (intentional or unintentional) misuse of the technology.
        \item If there are negative societal impacts, the authors could also discuss possible mitigation strategies (e.g., gated release of models, providing defenses in addition to attacks, mechanisms for monitoring misuse, mechanisms to monitor how a system learns from feedback over time, improving the efficiency and accessibility of ML).
    \end{itemize}

\item {\bf Safeguards}
    \item[] Question: Does the paper describe safeguards that have been put in place for responsible release of data or models that have a high risk for misuse (e.g., pre-trained language models, image generators, or scraped datasets)?
    \item[] Answer: \answerNA{}.
    \item[] Justification: Experiments are on purely synthetic data.
    \item[] Guidelines:
    \begin{itemize}
        \item The answer \answerNA{} means that the paper poses no such risks.
        \item Released models that have a high risk for misuse or dual-use should be released with necessary safeguards to allow for controlled use of the model, for example by requiring that users adhere to usage guidelines or restrictions to access the model or implementing safety filters.
        \item Datasets that have been scraped from the Internet could pose safety risks. The authors should describe how they avoided releasing unsafe images.
        \item We recognize that providing effective safeguards is challenging, and many papers do not require this, but we encourage authors to take this into account and make a best faith effort.
    \end{itemize}

\item {\bf Licenses for existing assets}
    \item[] Question: Are the creators or original owners of assets (e.g., code, data, models), used in the paper, properly credited and are the license and terms of use explicitly mentioned and properly respected?
    \item[] Answer: \answerYes{}
    \item[] Justification: The baselines of \citet{Reyd2025} (actualcauses, MIT licence) and \citet{Ibrahim2020} (HP2SAT1.0, MIT licence) are cited in the paper and used unchanged under their licences; their code is referenced in our repository. All other code was written by the authors with the help of coding agents.
    \item[] Guidelines:
    \begin{itemize}
        \item The answer \answerNA{} means that the paper does not use existing assets.
        \item The authors should cite the original paper that produced the code package or dataset.
        \item The authors should state which version of the asset is used and, if possible, include a URL.
        \item The name of the license (e.g., CC-BY 4.0) should be included for each asset.
        \item For scraped data from a particular source (e.g., website), the copyright and terms of service of that source should be provided.
        \item If assets are released, the license, copyright information, and terms of use in the package should be provided. For popular datasets, \url{paperswithcode.com/datasets} has curated licenses for some datasets. Their licensing guide can help determine the license of a dataset.
        \item For existing datasets that are re-packaged, both the original license and the license of the derived asset (if it has changed) should be provided.
        \item If this information is not available online, the authors are encouraged to reach out to the asset's creators.
    \end{itemize}

\item {\bf New assets}
    \item[] Question: Are new assets introduced in the paper well documented and is the documentation provided alongside the assets?
    \item[] Answer: \answerYes{}
    \item[] Justification: The structure and license of the accompanying code repository, as well as the required commands for running experiments and plotting results are described in the README of the repository. All code is released under MIT license.
    \item[] Guidelines:
    \begin{itemize}
        \item The answer \answerNA{} means that the paper does not release new assets.
        \item Researchers should communicate the details of the dataset\slash code\slash model as part of their submissions via structured templates. This includes details about training, license, limitations, etc.
        \item The paper should discuss whether and how consent was obtained from people whose asset is used.
        \item At submission time, remember to anonymize your assets (if applicable). You can either create an anonymized URL or include an anonymized zip file.
    \end{itemize}

\item {\bf Crowdsourcing and research with human subjects}
    \item[] Question: For crowdsourcing experiments and research with human subjects, does the paper include the full text of instructions given to participants and screenshots, if applicable, as well as details about compensation (if any)?
    \item[] Answer: \answerNA{}
    \item[] Justification: The paper does not involve crowdsourcing nor research with human subjects.
    \item[] Guidelines:
    \begin{itemize}
        \item The answer \answerNA{} means that the paper does not involve crowdsourcing nor research with human subjects.
        \item Including this information in the supplemental material is fine, but if the main contribution of the paper involves human subjects, then as much detail as possible should be included in the main paper.
        \item According to the NeurIPS Code of Ethics, workers involved in data collection, curation, or other labor should be paid at least the minimum wage in the country of the data collector.
    \end{itemize}

\item {\bf Institutional review board (IRB) approvals or equivalent for research with human subjects}
    \item[] Question: Does the paper describe potential risks incurred by study participants, whether such risks were disclosed to the subjects, and whether Institutional Review Board (IRB) approvals (or an equivalent approval/review based on the requirements of your country or institution) were obtained?
    \item[] Answer: \answerNA{}
    \item[] Justification: The paper does not involve crowdsourcing nor research with human subjects.
    \item[] Guidelines:
    \begin{itemize}
        \item The answer \answerNA{} means that the paper does not involve crowdsourcing nor research with human subjects.
        \item Depending on the country in which research is conducted, IRB approval (or equivalent) may be required for any human subjects research. If you obtained IRB approval, you should clearly state this in the paper.
        \item We recognize that the procedures for this may vary significantly between institutions and locations, and we expect authors to adhere to the NeurIPS Code of Ethics and the guidelines for their institution.
        \item For initial submissions, do not include any information that would break anonymity (if applicable), such as the institution conducting the review.
    \end{itemize}

\item {\bf Declaration of LLM usage}
    \item[] Question: Does the paper describe the usage of LLMs if it is an important, original, or non-standard component of the core methods in this research? Note that if the LLM is used only for writing, editing, or formatting purposes and does \emph{not} impact the core methodology, scientific rigor, or originality of the research, declaration is not required.
    \item[] Answer: \answerNA{}
    \item[] Justification: LLMs have only been used for writing, editing, and formatting of the paper and code in a way that does not impact the originality, rigor or core methodology of the paper.
    \item[] Guidelines:
    \begin{itemize}
        \item The answer \answerNA{} means that the core method development in this research does not involve LLMs as any important, original, or non-standard components.
        \item Please refer to our LLM policy in the NeurIPS handbook for what should or should not be described.
    \end{itemize}

\end{enumerate}

\end{document}